\documentclass{ieeeaccess}
\usepackage{cite}
\usepackage{amsmath,amssymb,amsfonts}
\usepackage{algorithmic}
\usepackage{tensor}
\usepackage{physics}
\usepackage{graphicx}
\usepackage{textcomp}
\usepackage{caption}
\usepackage{hyperref}
\usepackage{subcaption}
\usepackage{multirow}
\usepackage{comment}
\usepackage[table]{xcolor}
\definecolor{Gray}{gray}{0.85}

\makeatletter
\def\hlinewd#1{%
\noalign{\ifnum0=`}\fi\hrule \@height #1 \futurelet
\reserved@a\@xhline}
\makeatother

\def\BibTeX{{\rm B\kern-.05em{\sc i\kern-.025em b}\kern-.08em
    T\kern-.1667em\lower.7ex\hbox{E}\kern-.125emX}}
\begin{document}
\history{Date of publication xxxx 00, 0000, date of current version xxxx 00, 0000.}
\doi{10.1109/ACCESS.2017.DOI}

\title{Dynamic Based Estimator for UAVs with Real-time Identification Using DNN and the Modified Relay Feedback Test}
\author{\uppercase{Mohamad Wahbah}\href{https://orcid.org/0000-0003-1647-8546}{\includegraphics[scale=0.75]{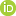}}\authorrefmark{1,*},
\uppercase{Mohamad Chehadeh}\href{https://orcid.org/0000-0002-9430-3349}{\includegraphics[scale=0.75]{orcid.png}}\authorrefmark{1,*},\IEEEmembership{Member, IEEE} and \uppercase{Yahya Zweiri}\href{https://orcid.org/0000-0003-4331-7254}{\includegraphics[scale=0.75]{orcid.png}}.\authorrefmark{1,2},
\IEEEmembership{Member, IEEE}}
\address[1]{Khalifa University Center for Autonomous Robotic  Systems, Khalifa University, Abu Dhabi, United Arab Emirates}
\address[2]{Faculty of Science, Engineering and Computing, Kingston University London, London SW15 3DW, U.K.}
\address[*]{These authors have contributed equally to this work}

\tfootnote{This work was supported by the Khalifa University of Science and Technology under Awards RC1-2018-KUCARS and CIRA-2020-082.}

\markboth
{Author \headeretal: Preparation of Papers for IEEE TRANSACTIONS and JOURNALS}
{Author \headeretal: Preparation of Papers for IEEE TRANSACTIONS and JOURNALS}

\corresp{Corresponding author: Mohamad Wahbah (e-mail: mohamad.wahbah@ku.ac.ae).}

\begin{abstract}
Control performance of Unmanned Aerial Vehicles (UAVs) is directly affected by their ability to  estimate their states accurately. With the increasing popularity of autonomous UAV solutions in real world applications, it is imperative to develop robust adaptive estimators that can ameliorate sensor noises in low-cost UAVs. Utilizing the knowledge of UAV dynamics in estimation can provide significant advantages, but remains challenging due to the complex and expensive pre-flight experiments required to obtain UAV dynamic parameters. In this paper, we propose two decoupled dynamic model based Extended Kalman Filters for UAVs, that provide high rate estimates for position, and velocity of rotational and translational states, as well as filtered inertial acceleration. The dynamic model parameters are estimated online using the Deep Neural Network and Modified Relay Feedback Test (DNN-MRFT) framework, without requiring any prior knowledge of the UAV physical parameters. The designed filters with real-time identified process model parameters are tested experimentally and showed two advantages. Firstly, smooth and lag-free estimates of the UAV rotational speed and inertial acceleration are obtained, and used to improve the closed loop system performance, reducing the controller action by over \(6\%\). Secondly, the proposed approach enabled the UAV to track aggressive trajectories with low rate position measurements, a task usually infeasible under those conditions. The experimental data shows that we achieved estimation performance matching other methods that requires full knowledge of the UAV parameters.
\end{abstract}


\titlepgskip=-15pt

\maketitle

\section*{Nomenclature}
\addcontentsline{toc}{section}{Nomenclature}
\begin{IEEEdescription}[\IEEEusemathlabelsep\IEEEsetlabelwidth{$V_1,V_2,V_3$}]
\item[\(\mathcal{W}\)] Inertial Frame with \(\mathcal{W} = \{\vb*{w_x}, \vb*{w_y}, \vb*{w_z}\}\)
\item[\(\mathcal{B}\)] Body Fixed Frame with \(\mathcal{B} = \{\vb*{b_x}, \vb*{b_y}, \vb*{b_z}\}\)
\item[\(\mathcal{S}\)] Sensor Fixed Frame with \(\mathcal{S} = \{\vb*{s_x}, \vb*{s_y}, \vb*{s_z}\}\)
\item[\({^\mathcal{F}}{\vb*{v}}{}\)] A vector \(\vb*{v}\) described in frame \(\mathcal{F}\), with basis \(\{\vb*{v_x}, \vb*{v_y}, \vb*{v_z}\}\)
\item[\({^\mathcal{F}}{\vb*{\bar{v}}}{}\)] A 2-D vector projection of vector \(\vb*{v}\) on the basis forming the x-y plane of frame \(\mathcal{F}\)
\item[\({^\mathcal{W} _\mathcal{B}}{R}{}\)] An ortho-normal rotation matrix that describes the orientation of frame \(\mathcal{B}\) with respect to frame \(\mathcal{W}\),  \({^\mathcal{B} _\mathcal{W}}{R}={^\mathcal{W} _\mathcal{B}}{R}^{-1}={^\mathcal{W} _\mathcal{B}}{R}^{T}\)
\item[\(I\)] Diagonal matrix consisting of the UAVs inertia \(diag(I_x, I_y, I_z)\) around the \(\mathcal{B}\) principal axes.
\item[\(\vb*{q}\)] A quaternion defined as \(\vb*{q} = \{q_w, q_x, q_y, q_z\}\)
\item[\(\theta\)] The angle of rotation around \(\vb*{w_x}\), referred to as the roll of the UAV
\item[\(\phi\)] The angle of rotation around \(\vb*{w_y}\), referred to as the pitch of the UAV
\item[\(\psi\)] The angle of rotation around \(\vb*{w_z}\), referred to as the yaw of the UAV
\item[\(\vb*{\omega}\)] Rotational velocity of the UAV
\item[\(\vb*{\alpha}\)] Rotational acceleration of the UAV
\item[\(\vb*{p}\)] Position of the UAV
\item[\(\vb*{v}\)] Linear velocity of the UAV
\item[\(\vb*{a}\)] Linear acceleration of the UAV
\item[\(\vb*{a_{bs}}\)] Body specific accelererations of the UAV
\item[\(m\)] Mass of the UAV
\item[\(\vb*{g}\)] The gravity field vector
\item[\(k_T\)] Thrust coefficient of the rotor
\item[\(k_C\)] Motor command to rotational velocity coefficient
\item[\(u\)] Motor command sent to the electronic speed controllers
\item[\(\mu_i\)] Is the rotational velocity of the \(i^{th}\) rotor
\item[\(\vb*{\Lambda}\)] Vector of translational drag coefficients in the \(\mathcal{B}\) frame; \(\vb*{\Lambda} = [\lambda_x \; \lambda_y \; \lambda_z]\)
\item[\(\vb*{\Gamma}\)] Vector of rotational drag coefficients in the \(\mathcal{B}\) frame; \(\vb*{\Gamma} = [\gamma_{x} \; \gamma_{y}]\)
\item[\(K_{{eq}}\)] The equivalent gain of the transfer function accounting for propulsion gain and inertia
\item[\(\tau\)] Delay of the system
\item[\(T_{prop}\)] Is the time constant of the propulsion system
\item[\(T_{\lambda_i}\)] Is the translation time constant of the body, due to drag, along the axis \(i\)
\item[\(T_{\gamma_i}\)] Is the rotation time constant of the body, due to drag, around the axis \(i\)
\item[\(T_{\Lambda}\)] Is a diagonal matrix of the inertial time constants, \(T_{\Lambda}\) = \(diag(T_{\lambda_x}, T_{\lambda_y}, T_{\lambda_z})\)
\item[\(T_{\Gamma}\)] Is a diagonal matrix of the rotational time constants, \(T_{\Gamma}\) = \(diag(T_{\gamma_x}, T_{\gamma_y})\)
\item[\(u_T\)] Collective motor commands sent to all electronic speed controllers generating thrust force along \(\vb*{b_z}\)
\item[\(\vb*{\bar{u}_M}\)] Differential motor commands sent to opposing electronic speed controllers causing a moment around \(\vb*{b_x}\) and \(\vb*{b_y}\) respectively
\item[\(\beta\)] The Modified Relay Feedback Test phase parameter
\item[\(h\)] The Modified Relay Feedback Test amplitude
\end{IEEEdescription}

\section{Introduction}
\label{sec:introduction}
\PARstart{U}{manned} Ariel Vehicles (UAVs) increasing popularity can be attributed to their versatility and high maneuverability compared to other robotic platforms. Such traits make them an attractive option for myriad of industries, including defence, agriculture, and entertainment. Autonomy in such applications is highly desired, due to economical, operational, and even security constraints. High bandwidth state measurements are essential for a satisfactory performance of such autonomous robotic systems. As the demand for autonomous UAVs increases, the per unit cost of UAVs need to be kept minimal and the development of new algorithms to leverage the full on-board capabilities is needed.
\subsection{Relevant Work}
Most commonly, a UAV will have an Inertial Measurement Unit (IMU) on-board. IMU's usually consists of three axes gyroscopes, accelerometers, and magnetometers. The literature is mature with a variety of kinematic estimators that estimate UAV orientation from IMU measurements \cite{Mahony2008,Bonnabel2009,Sharman2020}. The shortcoming of these filters is that they assume quasi-stationary flight for certain assumptions on stability and performance to hold. These orientation estimators are usually augmented with position measurements from external sources, such as GPS, Motion Capture (MoCap), Radio Frequency Identification (RFID) \cite{Zhang2019}, and Ultra Wide Bandwith (UWB) \cite{Weide2020}), or from on-board measurements (e.g. visual odometry \cite{Sharman2018}) to achieve several advantages. Some of these advantages are achieving higher update rate of position, estimating velocity, correcting for IMU biases, and providing a phase lead in position and velocity estimates as delays in IMU measurements are negligible.

Another class of filters use kinetics, in addition to the kinematic equations of motion, and thus we refer to them as dynamic filters. Such dynamic filters utilize model knowledge and can be used to provide smoother state estimates, to estimate additional states, or to estimate physical properties. There are vast differences in the structure of these filters. A notable contribution made by \cite{Martin2010} explains how an accelerometer measures body specific accelerations only. These accelerations are caused by forces that affect the housing of the accelerometer, but not the sensor itself. In other words the accelerometer measures the generated thrust, the induced drag, and other forces acting on the quadrotor body, but not the gravity. The work of \cite{Mahony2012} utilized this fact to develop a tunable deterministic observer that also uses feedforward drag estimates for platform velocity estimation. In \cite{Abeywardena2013}, the authors developed an Extended Kalman Filter (EKF) that estimates attitude angles, inertial lateral velocities, and gyroscope biases. The developed EKF utilized the under-actuation constraints that govern multirotor UAV dynamics which helped in reducing velocity drift. The work of \cite{Leishman2014} showed similar performance to \cite{Abeywardena2013} but had the advantage of predicting the drag parameters online.
Predicting the drag parameters online was possible due to the fact that these parameters were observable \cite{Leishman2014}.

The designed filters in \cite{Mahony2012,Abeywardena2013,Leishman2014} did not incorporate motors' thrust or motors' command in the state estimation. The work of \cite{Yong2019} extended the work of \cite{Leishman2014} to provide low-drift state estimates in three dimensions by incorporating pulse width modulation (PWM) motor commands sent to the electronic speed controllers (ESC). In \cite{Yong2019} a static map between motor's command and motor's thrust was found offline and utilized in the EKF and unscented Kalman filter (UKF) estimators. Such mapping completely disregards motor dynamics which will result in poor estimate of high frequency components. More recently, \cite{Svacha2019} developed a UKF that utilized motor speed measurements to obtain motor's thrust. With the availability of motor's thrust measurement (through motor's rotational speed measurements) it was possible to estimate the lateral and vertical drag terms, the inertia of the UAV, the inertia of the blades, the motor torque constant, the mass of the UAV, along with its attitude and velocity. The drawback of such approach is the absence of motor's rotational speed from most commercially available UAVs.

None of the above filters provided smooth estimate for inertial accelerations which can be quite useful for high performance feedback control. Estimates of inertial accelerations might seem redundant, as body accelerations are measured directly, and customarily at a high rate; however these measurements carry noises and biases that render them unusable in their raw form. The authors in \cite{Tal2021} utilized accurate inertial acceleration measurements and thrust estimates from motor's rotational speed measurements to perform aggressive maneuvers. The acceleration measurements from the IMU in \cite{Tal2021} are low-pass filtered to reduce the noise. The usage of such linear filters lags the acceleration measurement which can degrade control performance. Thus \cite{Hamandi2020} proposed an adaptive nonlinear filtering technique based on a manually tuned empirical signal and noise models that does not introduce lag to the acceleration signal. Adaptability of such nonlinear manually tuned filtration techniques to different UAV setups can be limited due to the variety of noise sources.
\subsection{Contributions}
The estimation of high rate and lag-free inertial acceleration, and rotational speed of UAV body in the absence of high-end IMUs and motor's rotational speed measurements motivates the contributions of this paper. Such sensory setup dominates the current market of commercial UAVs. Moreover, expensive calibration setups to estimate motor dynamics, or drag terms are often required with existing filters in literature. These two reasons justify the importance of the contribution from the practical and commercial aspects. Thus we propose a decoupled dynamics-based approach for the estimation of rotational and translational states. The rotational dynamic based Extended Kalman Filter (RDEKF) uses the attitude control commands as inputs, and the attitude estimates and the gyro signals as measurements, to estimate the rotational velocities. While the translational dynamic based Extended Kalman Filter (TDEKF) uses the thrust command, the attitude estimates, the accelerometer measurements, and position measurements to estimate the inertial acceleration, velocity and position of the UAV. 

The contributions of this paper are as follow, first we develop the capabilities of the DNN-MRFT algorithm proposed in \cite{Ayyad2020} for estimator tuning. This is done by providing the non-linear UAV model adopted by the majority of literature, and verifying that the proposed system model can capture the relevant dynamic trends. We chose this method of parameter estimations as it has the advantages of being online, light weight, easy to deploy, and requires no prior knowledge of any of the UAV parameters. The DNN-MRFT can be performed using off-the-shelf on-board sensors for the case of the RDEKF, and a wide variety of position sensors for the case of the TDEKF. Additionally, it accounts for actuator dynamics in the identification thus the assumption of a static relationship between motor's command and thrust is not required. The second contribution is the inclusion of rotational speeds and translational accelerations as estimates in the RDEKF and TDEKF. We show that the RDEKF and TDEKF with the estimated parameters provide significantly smoother and lag-free estimates of the rotational speeds and inertial accelerations, achieving an average Root Mean Square Error (RMSE) of \(0.1532 \; m/s^2\) and \(0.0459 \; rad/s\) respectively. These results represent a significant improvement compared to raw gyro and accelerometer measurements. It follows that the RDEKF and TDEKF estimates resulted in lower energy of the control signal, reducing the controller action by approximately \(6.6\%\). We validate the results experimentally by accurately tracking a figure-eight maneuver with low rate position measurements. Such maneuvers are usually infeasible without high rate measurements. However by using the filters estimates as control variables, the maneuver was executed successfully. Thus showing that our proposed approach can enable accurate tracking with low frequency position measurements systems, such as on-board cameras, or a Real Time Kinematic (RTK) system.
\subsection{Structure of the Paper}
This paper is organized as follows. In Section \ref{sec:sysmodel} we go over the nonlinear dynamic model presented in literature, and use it to obtain a decoupled model that assumes linear kinetics and nonlinear kinematics. Then in Section \ref{sec:ddekf} we cover the design of the RDEKF and TDEKF based on the developed decoupled nonlinear model. In Section \ref{sec:dnnmrft} we briefly describe how the dynamic model parameters can be obtained using the DNN-MRFT approach. experimental results are shown in Section \ref{sec:res} highlighting the advantages offered by the proposed approach. Finally, Section \ref{sec:conc} concludes the article with a discussion and an outlook. 
\begin{figure} [t]
    \centering
    \includegraphics[width=7.5 cm]{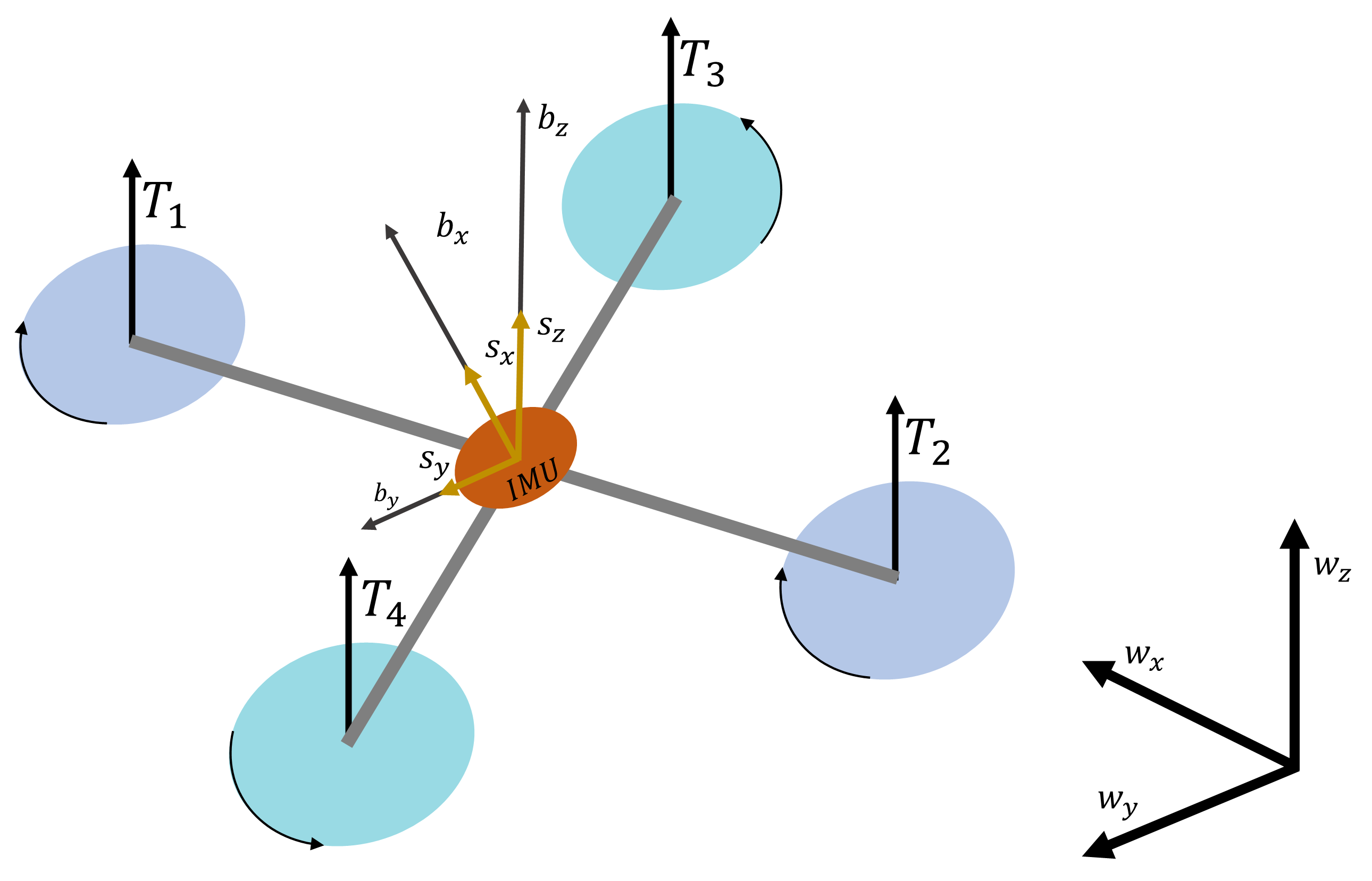}
    \caption{Notation and reference frames used in this paper. Without loss of generality, \(\mathcal{B}\) is defined assuming "X" quadrotor configuration.}
    \label{fig:uavmodel}
\end{figure}
\section{System Model} \label{sec:sysmodel}
\subsection{Nonlinear Model}
As accelerations are caused by forces that affect the housing of the accelerometer and not the sensor itself, it is necessary to model all forces acting on the quadrotor body. This will relate the measurements obtained by the inertial sensors, and our proposed model. Hence we aim first to develop a nonlinear model that captures said forces, then we introduce a few assumptions and linearize parts of the model to make the real-time performance, and identification with DNN-MRFT feasible. The reference frames used in this work are defined in the Nomenclature and are shown in Fig. \ref{fig:uavmodel} for illustration.

For a body motion that abides to Newtonian dynamics, the summation of forces is:
\begin{equation}
    m\vb*{a} = \sum \vb*{F}
\end{equation}
where \(F\) is any force applied to the body. For a quadrotor, this can be expanded into:
\begin{equation}
\label{eq:forces}
    m\vb*{a} = \vb*{F_{t}} - \vb*{F_{d}} - \vb*{F_{g}}
\end{equation}
The total generated  thrust \(\vb*{F_{t}}\) is due to the rotors' rotations, and can be modeled as:
\begin{equation}
\label{eq:forces_thrust}
    {^\mathcal{B}}\vb*{F_{t}} = k_T \sum_{i = 1}^{4} \mu^2_i \vb*{b_z}
\end{equation}
where a rotor's rotation is related to the ESC input command by:
\begin{equation}
    \mu_i = k_C u_i
\end{equation}
where \(u_i\) is the motor command sent to the \(i^{th}\) motor. Note that in this paper we will also use \(u_{i}\) with \(i \in \{b_x,b_y,z\}\) to conveniently describe differential, or collective motor thrust commands that correspond to the outputs of the individual inner control loops, for rotations around \(b_x\) and \(b_y\), and collective thrust.

The induced translational drag forces \(\vb*{F_{d}}\) consists of two parts as seen from \(\mathcal{B}\) \cite{POUNDS2010,Hoffmann2007}. The first part of the drag is lateral (i.e. co-planar with \(\vb*{b_x}\cross\vb*{b_y}\)) and is mainly attributed to blade flapping, and profile drag on the rotors and the UAV body. The second part of the drag, which is the prominent one, is along \(\vb*{b_z}\) and is mainly caused by the change in inflow angle, and the profile drag on the rotors and the UAV body. The change in inflow angle changes the thrust coefficient \(k_T\), and was approximated in \cite{POUNDS2010} for the \(i^{th}\) motor to be:
\begin{equation}
    k_{T_i}=k_{T_0}+\Delta k_{T_i}
\end{equation}
where the change increment, \(\Delta k_{T_i}\), is given by:
\begin{equation}
\label{eq:motion_inflow_change}
    \Delta k_{T_i}= \frac{c_r}{\mu_i}{^\mathcal{B}}{\vb*{v}{^h_i}}
\end{equation}
where \(c_r\) is a constant of the rotor's physical properties, and \({^\mathcal{B}}{\vb*{v}{^h_i}}\) is the velocity of the \(i^{th}\) rotor hub, given by:
\begin{equation}
    {^\mathcal{B}}{\vb*{v}{^h_i}} = {^\mathcal{B}}{\vb*{v}}{} + {^\mathcal{B}}{\vb*{\omega}}{} \cross \vb*{r_i}
\end{equation}
where \(r_i\) is the position of the center of the \(i^{th}\) motor relative to the UAV center of mass. Thus, rotor motion inflow not only damps translations along \(\vb*{b_z}\), but also rotational motion by producing opposing moments. It follows that \(\vb*{F_{d}}\) can be modeled as:
\begin{equation}
    {}^{\mathcal{B}}\vb*{F_{d}} = ({}^{\mathcal{B}}\vb*{F_{m}}\cdot{\vb*{b_z}}){\vb*{b_z}} + \vb*{\Lambda_b} \odot {^\mathcal{B}}{\vb*{v}}
    \label{eq:forces_drag}
\end{equation}
where \(\odot\) is the Hadamard product, \({}^{\mathcal{B}}\vb*{F_{m}}\) is the rotor motion inflow drag due to rigid body translation, and \(\vb*{\Lambda_b}\) is a vector of profile drag coefficients. Finally, the gravity force is given by:
\begin{equation}
    {^\mathcal{B}}\vb*{F_{g}}=m{^\mathcal{B}}\vb*{g}
\end{equation}

On the other hand, the rotation dynamics of the UAV are given by:
\begin{equation} \label{eq:moments}
    I \frac{d}{dt} ({^\mathcal{B}}{\vb*{\omega}}{}) = {^\mathcal{B}}{\vb*{M}}{_t} -
    {^\mathcal{B}}{\vb*{M}}{_g} -
    {^\mathcal{B}}{\vb*{M}}{_d} -
    {^\mathcal{B}}{\vb*{M}}{_f}
\end{equation}
The moment vector \({^\mathcal{B}}{\vb*{M}}{_t}\), generated by the rotors due to rotation is \cite{Svacha2019}:
\begin{equation} \label{eq:moments_thrust}
    {^\mathcal{B}}{\vb*{M}}{_t} = \sum_{i = 1}^{4}\big( \vb*{r_i} \cross \vb*{F_i} - d(i) k_m \mu^2_i \vb*{b_z} \big)
\end{equation}
where \(\vb*{F_i}\) is the thrust force of the \(i^{th}\) motor, \(d(i)\) is a function of the motor direction, that returns one for counter-clock wise rotations, and negative one for clock wise rotations, and \(k_{m}\) is a positive coefficient. The other moments are due to the gyroscopic effect \({^\mathcal{B}}{\vb*{M}}{_g}\), the profile drag and motion inflow on UAV body and propellers \({^\mathcal{B}}{\vb*{M}}{_d}\), and due to blade flapping \({^\mathcal{B}}{\vb*{M}}{_f}\). The gyroscopic moments are given by:
\begin{equation}
\label{eq:moments_gyroscopic}
    {^\mathcal{B}}{\vb*{M}}{_g}={^\mathcal{B}}{\vb*{\omega}}{} \cross I \; {^\mathcal{B}}{\vb*{\omega}}{}
\end{equation}
And the moments caused by the profile drag and motion inflow are given by:
\begin{equation}
\label{eq:moments_drag}
    {^\mathcal{B}}{\vb*{M}}{_d}= {^\mathcal{B}}{\vb*{M}}{_m} + \vb*{\Lambda_b} \odot {^\mathcal{B}}{\vb*{w}}
\end{equation}
where \({^\mathcal{B}}{\vb*{M}}{_m}\) is a vector of the moments due to the motion inflow caused by angular rotation as presented in Eq. \eqref{eq:motion_inflow_change}. Lastly, the moments caused by blade flapping are given by:
\begin{equation}
    \label{eq:moments_flapping}
    {^\mathcal{B}}{\vb*{M}}{_f} = \sum_{i = 1}^{4} k_f \; \mu_i {^\mathcal{B}}{\vb*{v}}{^h_i} \cross \vb*{b_z}
\end{equation}
where \(k_f\) is a positive coefficient.
\subsection{Linearized and Loosely Coupled Models} \label{sec:linmodels}
Designing a dynamic estimator based on the form of nonlinear coupled equations presented in Eqs.\eqref{eq:forces} and \eqref{eq:moments} poses two challenges. The first challenge is associated with the computational and tuning complexity of a high dimensional strongly coupled EKF estimator. The second challenge is related to how can we identify or estimate the physical parameters of the UAV. We solve the first challenge by loosening the coupling between the rotational and translational dynamics. The only coupling that remains between the two is due to kinematics, which cannot be ignored. The second challenge is solved by performing accurate online identification of lumped system parameters through DNN-MRFT. We refer to them as lumped parameters due to the fact that a single parameter may be used to capture a few physical phenomena that have the same effect from the dynamics perspective. For example, drag from motion inflow and profile drag are described by one lumped parameter.

Based on the work proposed by \cite{Ayyad2020}, the vertical (i.e. the movement along \(\vb*{b_z}\)), and the pitch and roll control loops can be modeled as a second order with integrator plus time delay (SOIPTD) processes. While lateral motion is split into two cascaded processes, the inner process is the roll or pitch, and the outer first order with integrator plus time delay (FOIPTD) process is for the corresponding motion along \(\vb*{w_x}\) or \(\vb*{w_y}\). In this section, we go over elementary SISO models for the vertical, attitude, and lateral loops that are suitable for physical parameters identification. Then we present two unified models, one for translational motion, and another one for the rotational motion, that composes the identified SISO models with nonlinear kinematics to form the TDEKF and RDEKF state estimators.

Note that we have omitted the yaw moment as estimation of the yaw dynamics is not of interest in this work, where the focus is devoted to fast transient dynamics that are present in aggressive maneuvers. Moreover, control and estimation for yaw dynamics is simpler due to the presence of full state measurements and the lower relative degree of the system \cite{ayyad2021tcst}. Therefore in this work we are using a kinematic estimator for the yaw states based on \cite{Mahony2008}.

Let us first consider the SOIPTD model of the vertical motion dynamics:
\begin{equation}
\label{eq:siso_altitude}
G_{z}(s)=\frac{Z(s)}{U_T(s)} = \frac{K_{prop_z}}{(T_{prop_z}s + 1)} \cdot \frac{K_z}{(T_{\lambda_z} s + 1)} \cdot \frac{1}{s} \cdot e^{\tau_z s}
\end{equation}
this model consists of two first order systems, an integrator, and a time delay. The first of the two first order systems along with part of the time delay models the propulsion dynamics, while the second models the effective drag along the vertical motion. The drag parameter \(T_{\lambda_z}\) provides a lumped description of the profile and motion inflow drags presented in Eq. \eqref{eq:forces_drag} along \(\vb*{b_z}\). Linear drag models where found to be good approximates of the underlying physics even at translational speeds of few meters per second \cite{POUNDS2010,Torrente2021}. The rest of the time delay is used to capture the delay in the position measurement. The time delay is distributed among the forward (i.e. delay in propulsion) and feedback (i.e. delay in sensor) paths of the control loop. The DNN-MRFT is an input-output identification method, hence it cannot distinguish these delays. For the simplicity of computation we assume that all the delay is in the forward path. The propulsion gain \(K_{{prop}_z}\) is an approximation of the nonlinear propulsion equation shown in Eq. \eqref{eq:forces_thrust} around the hover point. Similar to the time-delay case, DNN-MRFT cannot distinguish \(K_{{prop}_z}\) from the inertia gain \(K_{z}\), and therefore we consider \(K_{{eq}_z}=K_{{prop}_z}K_{z}\). 

\begin{figure*} [t]
    \centering
    \includegraphics[width = \textwidth]{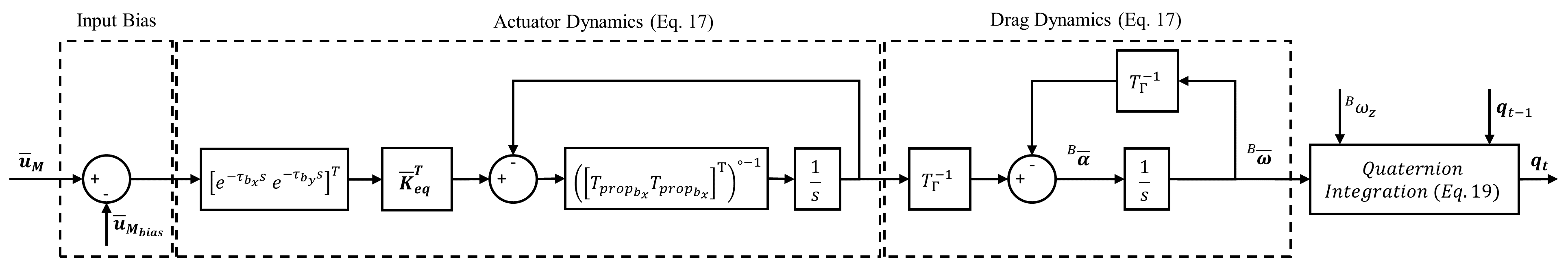}
    \caption{Estimator model used in the translational filter}
    \label{fig:TDEKFmodel}
\end{figure*}

The SISO linear model for the lateral motion, which relates the UAV attitude angle to the UAV lateral position is given by:
\begin{equation} \label{eq:siso_outer}
G_{i}(s)= \frac{X_i(s)}{\Theta_i(s)} = \frac{K_{i}}{(T_{\lambda_i^t} s + 1)} \cdot \frac{1}{s} \cdot e^{-\tau_{i} s}
\end{equation}
with \(i \in \{x,y\}\), and \(T_{\lambda_i}\) represents the time constants for the drag. This structure provides a lumped linear approximation of the profile drag components of Eq. \eqref{eq:forces_drag} along \(\vb*{b_x}\) and \(\vb*{b_y}\) and the blade flapping damping moments shown in Eq. \eqref{eq:moments_flapping}. The time delay in position measurement \(e^{\tau_{x, y} s}\) and the equivalent gain \(K_{x, y}\) are not used in the estimators, but they have essential role in the design of optimal feedback controllers \cite{ayyad2021tcst}. Each of the translational drag time constants is used to form the drag matrix \(T_{\Lambda}\). The overall TDEKF translational dynamics are shown in Fig. \ref{fig:TDEKFmodel} which uses the lumped system parameters from Eqs. \eqref{eq:siso_altitude} and \eqref{eq:siso_outer}, and the nonlinear kinematics of multirotor UAVs.

\begin{figure*} [t]
    \includegraphics[width = \textwidth]{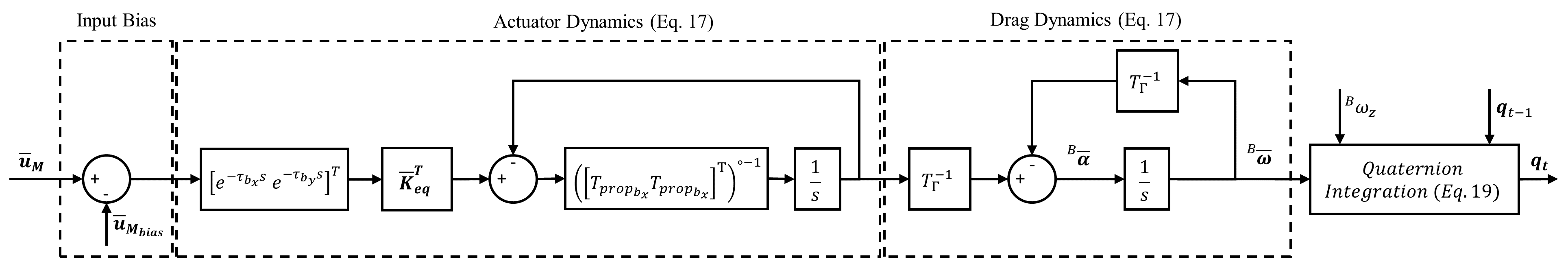}
    \caption{Estimator model used in the rotational filter}
    \label{fig:RDEKFmodel}
\end{figure*}

The attitude loops have the same structure of the altitude loop and they are modeled as:
\begin{equation}
\label{eq:siso_attitude}
G_{i}(s) = \frac{\Theta_i(s)}{U_i(s)} = \frac{K_{prop_{i}}}{(T_{prop_{i}}s + 1)} \cdot \frac{K_{i}}{(T_{\gamma_j} s + 1)} \cdot \frac{1}{s} \cdot e^{\tau_{i} s}
\end{equation}
with \(i \in \{b_x, b_y\}\), and \(j \in \{x, y\}\). The lumped drag time constants \(T_{\gamma_i}\) represent the damping due to rotor motion inflow and profile drag moments as described in Eq. \eqref{eq:moments_drag}. Note that the gyroscopic moments presented in Eq. \eqref{eq:moments_gyroscopic} are ignored in this formulation to remove the coupling between loops. It is argued in \cite{CHEHADEH2019} that the effect of the gyroscopic moments can be neglected for the range of the physical parameters of the common multirotor UAV sizes and design. The overall structure of the RDEKF is shown in Fig. \ref{fig:RDEKFmodel}, where quaternions are used to properly handle rotational speed integration through the use of the following equations:
\begin{equation}
    \dot{\vb*{q}} = \frac{1}{2}\vb*{q} \otimes \left[0, \; {^\mathcal{B}}{\omega}{_x}, \; {^\mathcal{B}}{\omega}{_y}, \; {^\mathcal{B}}{\omega}{_z} \right] 
\end{equation}
\begin{equation}
    \vb*{q}_{t} = \vb*{q}_{t-1} \oplus \; \dot{\vb*{q}} \Delta t
\end{equation}
where \(\vb*{q}_t\) is the new estimate of the orientation, \(\vb*{q}_{t-1}\) is the previous estimate, and \(\Delta t\) is the integration time step. The translational model uses the rotation matrix representation, hence we convert between the two representations using:
\begin{equation}
    {^\mathcal{W} _\mathcal{B}}{R}{} = 
    \begin{bmatrix}
        1-2q_y^2-2q_z^2      & 2(q_x q_y - q_w q_z) & 2(q_x q_z + q_w q_y) \\
        2(q_x q_y + q_w q_z) & 1-2q_x^2-2q_z^2      & 2(q_y q_z - q_w q_x) \\
        2(q_x q_z - q_w q_y) & 2(q_y q_z + q_w q_x) & 1-2q_x^2-2q_y^2
    \end{bmatrix}
\end{equation}

\section{Decoupled Dynamic Based Kalman Filter} \label{sec:ddekf}
\subsection{Estimator Design}
Let's donate \(\vb*{x_R}\) for the estimates vector of the RDEKF, and \(\vb*{x_T}\) for the estimates of the TDEKF. The vector \(\vb*{x_R}\) is  defined in \(\mathbb{R}^{13 \times 1}\), and consists of:
\begin{equation}
    \vb*{x_R} = \left[\; \;{\vb*{q}}\; \; {^\mathcal{B}}{\vb*{\bar{\omega}}}{} \; \; 
    {^\mathcal{B}}{\vb*{\bar{\alpha}}}{} \; \;  
    {^\mathcal{B}}{\vb*{\bar{\alpha_{t}}}}{} \; \; 
    {\vb*{\bar{u}_{M_{bias}}}} \; \; \right]
\end{equation}
where, \({^\mathcal{B}}{\vb*{\alpha_{t}}}{}\) is the inertia normalized thrust generated moment vector:
\begin{equation}
    {^\mathcal{B}}{\vb*{\alpha_{t}}}{} = \bar{\vb*{M}} \cdot \left[\; I_x^{-1} \; I_y^{-1} \; \right]^T
\end{equation}
and \(\vb*{\bar{u}_{M_{bias}}}\) is the process bias. This bias is used to offset rotations caused by motor mismatch. The states in \(\vb*{x_R}\), apart from \(\vb*{q}\), are all defined as 2-D vectors, for rotations around \(\vb*{b_x}\) and \(\vb*{b_y}\). The angle \(\psi\) is assumed to be measured externally, and \(\vb*{q}\) is updated accordingly.

\begin{figure*}
    \centering
    \includegraphics[width=\textwidth]{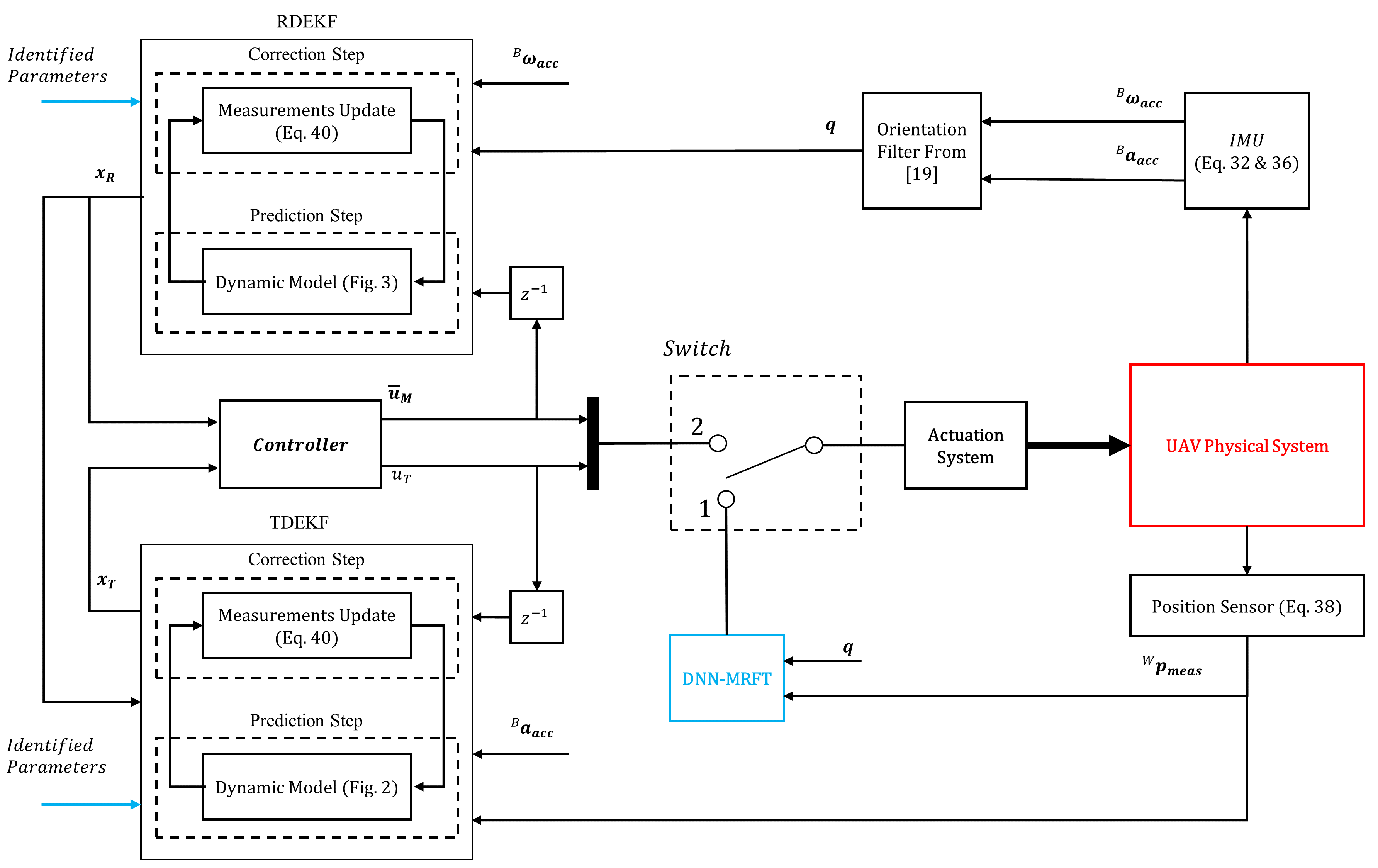}
    \caption{Structure of the proposed filtering scheme. The switch is at position 1 initially, initiating the DNN-MRFT algorithm. After the parameters have been identified, the RDEKF and TDEKF are models updated and the switch is flipped to position 2}
    \label{fig:schematics}
\end{figure*}

The TDEKF estimates vector \(\vb*{x_T}\) is  defined in \(\mathbb{R}^{11 \times 1}\) and consists of:
\begin{equation}
    \vb*{x_T} = \left[\; \;{^\mathcal{W}}{\vb*{p}}{} \; \; {^\mathcal{W}}{\vb*{v}}{} \; \; {^\mathcal{B}}{\vb*{a}}{_{bs}} \; \; 
    a_{T} \; \; 
    u_{T_{bias}} \; \; \right]
\end{equation}
where \(a_{T}\) is the mass normalized generated thrust:
\begin{equation}
    a_T = \frac{F_t}{m} \vb*{b_z}
\end{equation}
and \(u_{T_b}\) is the process bias. The bias here compensates for the slow drift in thrust command caused by battery voltage drop. These two states are modeled as scalar quantities rather than a vector as the generated thrust is always aligned with \(b_z\). The rest of the estimates in \(\vb*{x_T}\) are three dimensional quantities. Fig \ref{fig:schematics} shows the proposed filtering approach. Initially, the model parameters are not identified, and the mode is set to the DNN-MRFT. This will initiate the identification phase, and the UAV will excite oscillations as described in sec. \ref{sec:dnnmrft}. After the identification is completed and the models parameters are updated, the DNN-MRFT is disengaged, and the filters provide estimates to the actuation system instead. The RDEKF and TDEKF computes the estimates using a prediction step, and a measurement step. The prediction step uses the previous controller command to estimate the UAV current states. At the arrival of a new measurement, the measurement step is carried out to correct the predictions. In the following sections, we detail how these steps are implemented, and provide the measurement models used.
\subsection{Prediction Model}
The TDEKF and RDEKF process models presented in Figs. \ref{fig:TDEKFmodel} and \ref{fig:RDEKFmodel} are not linear due to the considered rotation kinematics. Hence we consider the use of EKF to linearize the process dynamics. For brevity, the full closed-form model and the linearization steps are not included in this document. A detailed workbook with the full estimator model derivations and its functions is provided in \cite{wahbah2021}.

The state-space model for an arbitrary nonlinear system with the states \(\vb*{x}(t)\), an input \(\vb*{u}(t)\), and additive process and measurement noises, is given by:
\begin{equation}
    \vb*{\dot{x}}(t) = f(\vb*{x}(t), \vb*{u}(t)) + \mathcal{N}(0, R_{proc})
\end{equation}
\begin{equation}
    \vb*{y}(t) = h(\vb*{x}(t), \vb*{u}(t))  + \mathcal{N}(0, R_{meas})
\end{equation}
where \(\mathcal{N}(0, R)\) is a zero mean normally distributed noise. The process noise variance is \(R_{proc}\), and the measurement noise variance is \(R_{meas}\). If we linearize the state equation \(f\) and the output equation \(h\) around the current states \(\vb*{x}\) and time \(t\), we can re-write the model as:
\begin{equation}
    \vb*{\dot{x}}(t) = A \vb*{x}(t) + B \vb*{u}(t) + \mathcal{N}(0, R_{proc})
\end{equation}
\begin{equation}
    \vb*{y}(t) = C \vb*{x}(t) +  D \vb*{u}(t) + \mathcal{N}(0, R_{meas})
\end{equation}
where \(A\) is the state matrix, \(B\) is the input matrix, \(C\) is the output matrix, and \(D\) is the feedforward matrix. In the case of the RDEKF, \(\vb*{u}(t)\) consists of \(u_{b_x}(t)\) and \(u_{b_y}(t)\), while for the TDEKF \(\vb*{u}(t)\) is \(u_T(t)\), and the orientation represented in \({^\mathcal{W} _\mathcal{B}}{R}{}\) at time \(t\).

For the discrete implementation, the state matrix is replaced by the transition matrix, which is defined as:
\begin{equation}
    F_k = e^{A \Delta t}
\end{equation}
where \(\Delta t\) is the filter prediction time step. Calculating the exponential of a matrix can be computationally expensive, in this work, we use a truncated Taylor series to linearize the state equation and find \(F_k\) for every \(t = k \Delta t \). Similarly, the control matrix is derived from the input matrix:
\begin{equation}
    G_k = \int_{0}^{\Delta t} e^{A(t) \Delta t} \, dt B(t)
\end{equation}
and is also found at every iteration using a truncated Taylor series. Finally, the prediction update step in discrete time is:
\begin{equation}
    \Tilde{\vb*{x}}_{k+1} = F_k \vb*{x}_k + G_k \vb*{u}_k
\end{equation}
where \((\Tilde{.})\) indicates the predicted quantity.
\subsection{Measurement Model}
As customary with these type of filters, the RDEKF and TDEKF require measurements from various sensors to correct the estimate from the prediction model. The TDEKF is updated with a 3-axis accelerometer measurements to compensate for the bias in thrust, and linearization errors, while position measurements provide correction for drift caused by biases in the predicted acceleration. The RDEKF uses 3-axis gyroscope measurements to correct for thrust bias, and orientation measurements to correct for angle drift. The need for orientation measurements arise from the decoupled nature of the filter. In this work we utilized the orientation estimation approach presented in \cite{Madgwick2011} to provide angle correction. Finally, since the inertial sensors sampling frequencies are usually much higher from that of a position sensor, the multi-rate sampling approach described in \cite{Quan2017} is used.

To incorporate the measurements in the filters we need to provide measurement models that relates the measured quantities, to the filter estimates. The gyroscope measurements model adopted in this work is:
\begin{equation}
    ^{\mathcal{B}}{\vb*{\omega}}_{gyro} = ^{\mathcal{B}}{\vb*{\omega}} + \mathcal{N}(0, R_{gyro})
\end{equation}
where \(R_{gyro}\) is the additive noise variance. The gyroscope directly measures the rotational velocity of the UAV, thus its observation matrix is:
\begin{equation}
    H_{gyro} = \begin{bmatrix} 0_{2 \times 4} & \mathcal{I}_{2 \times 2} & 0_{2 \times 2} & 0_{2 \times 2} & 0_{2 \times 2} \end{bmatrix}
\end{equation}
The orientation measurements can be modeled to have an additive measurement noise with variance \(R_{ang}\):
\begin{equation}
    \vb*{q}_{orient} = \vb*{q} + \mathcal{N}(0, R_{ang})
\end{equation}
The orientation provided by the measurement directly relates to the quaternion estimate from the filter, thus the orientation observation matrix is trivial:
\begin{equation}
    H_{orient} = \begin{bmatrix} \mathcal{I}_{4 \times 4} & 0_{4 \times 2} & 0_{4 \times 2} & 0_{4 \times 2} & 0_{4 \times 2} \end{bmatrix}
\end{equation}
The accelerometer measurement model used in the TDEKF is:
\begin{equation}
    \begin{split}
        ^{\mathcal{B}}{\vb*{a}}_{acc} = & \frac{1}{m} \big( k_T \sum_{n = 1}^{4} \mu^2_n {\vb*{b_z}} - \; ({}^{\mathcal{B}}\vb*{F_{m}}\cdot{\vb*{b_z}}){\vb*{b_z}} - \vb*{\Lambda_b} \odot {^\mathcal{B}}{\vb*{v}} \big)\; \\ & + \; {^\mathcal{B}}{\vb*{b}}{_{acc}} \; + \; \mathcal{N}(0, R_{acc})
    \end{split}
\end{equation}
where \(R_{acc}\) is the measurement noise variance. The accelerometer gives a direct measurement of \({^\mathcal{B}}{\vb*{a}}{_{bs}}\), hence, the accelerometer observation matrix has a trivial derivation:
\begin{equation}
    H_{acc} = \begin{bmatrix} 0_{3 \times 3} & 0_{3 \times 3} & \mathcal{I}_{3 \times 3} & 0_{3 \times 1} & 0_{3 \times 1} \end{bmatrix}
\end{equation}
It should be noted that the accelerometer is assumed to be perfectly aligned with the body of the UAV. Hence it provides measurements of the UAV acceleration, without the need for any transform. Additionally, the accelerometer is assumed to be calibrated using six-point tumble method \cite{6pointtumble}, and that the bias \(b_{acc}\) has been eliminated. The advantage of six-point tumble calibration over a simple bias calibration is the removal of sensitivity and cross-gains, thus allowing the decoupling between the filters. The drift in these quantities was noticed to be slow, especially when compared to the drift in thrust commands due to voltage drop. The position measurement model depends on the sensor used. In the experimental part of this work, we use a MoCap system. We also simulated other positioning systems by reducing the update rate of the MoCap system to match typical arrangements usually found in deployed UAV solutions; e.g. a GPS receiver with centimeter level accuracy, or a UWB localization system. The measurement model for the aforementioned systems is:
\begin{equation}
    \vb*{p}_{pos} = {^\mathcal{W}}\vb*{p} + \mathcal{N}(0, R_{pos})
\end{equation}
where \(R_{pos}\) is the noise variance, and the observation matrix is
\begin{equation}
    H_{pos} = \begin{bmatrix} \mathcal{I}_{3 \times 3} & 0_{3 \times 3} & 0_{3 \times 3} & 0_{3 \times 1} & 0_{3 \times 1}  \end{bmatrix}
\end{equation}
At the arrival of a new observation, the states are updated using the new measurement and the predicted states:
\begin{equation}
    \Hat{\vb*{x}}_k = \Tilde{\vb*{x}}_k + K_k (z_k - H \Tilde{\vb*{x}}_k)
\end{equation}
where \((\Hat{.})\) is used to represent the corrected estimate, \(K_k\) is the kalman gain, \(z\) is the measurement from a sensor, and \(H\) is the observation matrix of that measurement. 
\section{Identification of Model Parameters Through DNN-MRFT} \label{sec:dnnmrft}
The DNN-MRFT approach was suggested in \cite{Ayyad2020} to perform real-time identification and near-optimal tuning of UAV control loops. DNN-MRFT excites a stable periodic motion in the system that reveals the unknown system dynamics. The periodic motion is excited using the modified relay feedback test (MRFT) \cite{boiko2012nonparametricbook}. Measured periodic system output is then fed to a deep neural network (DNN) which classifies the unknown process and provides the corresponding system parameters. In this section we provide a brief discussion on DNN-MRFT and how it is used to find the unknown process parameters necessary for the RDEKF and TDEKF prediction models.
\subsection{The Modified Relay Feedback Test} \label{sec:mrft}
The MRFT is an algorithm that excites a plant to generate self-sustained oscillations at a specific phase, denoted by \(\Psi\). The algorithm is used as a controller, and can be implemented using the following equation:
\begin{multline}\label{eq_mrft_algorithm}
u_M(t)=\\
\left\{
\begin{array}[r]{l l}
h\; &:\; e(t) \geq b_1\; \lor\; (e(t) > -b_2 \;\land\; u_M(t-) = \;\;\, h)\\
-h\; &:\; e(t) \leq -b_2 \;\lor\; (e(t) < b_1 \;\land\; u_M(t-) = -h)
\end{array}
\right.
\end{multline}
where \(b_1 = -\beta e_{min}\) and \(b_2 = \beta e_{max}\), and \(u_M(-t)\), \(e_{max}\), and \(e_{min}\) are the previous command, maximum error, and minimum error, respectively. The phase \(\Psi\) is defined as:
\begin{equation}
    \Psi = \arcsin(\beta)
\end{equation}
In the DNN-MRFT framework, the parameter \(\Psi\) is a design parameter. Given a Linear Time Invariant (LTI) model structure \(G(s)\) that defines a set of processes with parameters in a subspace \(D\), the distinguishing phase \(\Psi_d\) is defined as the phase of the MRFT that when applied to said processes, the self-sustained oscillations would carry enough distinct information to identify their corresponding set of parameters in \(D\). The value of \(\Psi_d\) is found based on the process of optimal design of tuning rules (for details of this process refer to \cite{boiko2012nonparametricbook}).
\subsection{Process Classification based on Deep Neural Network} \label{sec:dnn}
In the DNN-MRFT approach, the mapping between the MRFT induced oscillations and the corresponding process is handled by a deep neural network classifier. The discretized subspace \(\Bar{D}\) contains the \emph{key processes} in \(D\) adhering to the criteria:
\begin{equation}
    \exists \; \Bar{d} \in \Bar{D} \; s.t. \; J(\Bar{d}, d) < 10\% ; \; \forall \; d \in D
\end{equation}
where \(J(\Bar{d}, d)\) is the relative sensitivity function defined by \cite{Ayyad2020}:
\begin{equation}
    J(\Bar{d}, d) = \frac{Q(C^*_{\Bar{d}}, G_d) - Q(C^*_{d}, G_d)}{Q(C^*_{d}, G_d)} \cross 100\%
\end{equation}
where \(C^*_{\Bar{d}}\), \(C^*_{d}\) are the optimal controllers of the discretized process and the actual process respectively, \(G_d\) is the actual process, and \(Q(C, G)\) is the Integral Square Error (ISE) performance index of applying controller \(C\) on process \(G\), for any \(C\) and \(G\). This approach in classification would not identify the exact process parameters, but would estimate a similar process whose closed loop performance, i.e. the input to output relationship, is similar. Discretization is essential for real-time performance as optimal tuning can be found offline, and also significantly lowers the training time of the DNN. Furthermore, the limit on \(J(\Bar{d}, d)\) ensures that the loss due to discretization is minimal.

The features chosen for identification were the controller output, and the observed process variable of a single periodic cycle. The identification process starts when the MRFT is triggered. The algorithm monitors the frequency of the self-excited oscillations, and upon reaching a constant frequency, a cycle is extracted and pre-processed to comply with the input of the DNN. A controller cycle is defined to be bounded by two consecutive transitions from \(-h\) to \(+h\). The process variable cycle is extracted using the time period of the identified controller cycle. Processes in \(\bar{D}\) would produce periodic cycles with different periods, and in such case, the input is padded with a vector of zeros to match the input size of the DNN classifier, thus the DNN input size is chosen such that the process with the lowest frequency in \(\Bar{D}\) can be represented. Two sets of DNN's were used for identification, the first set contained a single network, and is used to identify the attiude, and the altitude process parameters, while the second set is used to identify the lateral processes. The number of networks in the second set depends on the number of classes in the first set. This was imposed by the composition approach described in \cite{ayyad2021tcst}. In this work, the first set is used to estimate the SOIPTD processes parameters in Eqs \eqref{eq:siso_altitude} and \eqref{eq:siso_attitude}, and the second set is used for the FOIPTD parameters shown in Eq. \eqref{eq:siso_outer}. The architecture used in both sets was identical, as it showed to provide the best classification accuracy while doing a sweep of the hyper-parameters of different DNN architectures.
\subsection{Identification Methodology}
During the identification stage, it is important to ensure that the self-excited oscillations do not excite the nonlinearities of the physical system. For example, consider the nonlinearity of the proposed TDEKF process presented in Fig. \ref{fig:TDEKFmodel}. Keeping \(\vb*{w_z}\) aligned with \(\vb*{b_z}\) during identification ensures negligible contribution of the rotation matrix nonlinearity, i.e. \({^\mathcal{W} _\mathcal{B}}{R} \approx{}{}{} \mathcal{I}_{3 \times 3}\). Similarly, the identification of the drag parameters \(T_{\lambda_x}\) and \(T_{\lambda_y}\), which are defined for movements along \(\vb*{b_x}\) and \(\vb*{b_y}\) respectively, requires the self-excited oscillations to induce lateral movements while keeping \(\vb*{w_z}\) aligned with \(\vb*{b_z}\). Obviously this is not possible due to the under-actuated nature of UAVs, and the UAV thrust vector must change its orientation to generate accelerations along \(\vb*{w_x}\) or \(\vb*{w_y}\). Consequently, during the DNN-MRFT phase, the UAV will change its orientation slightly to produce oscillations during lateral processes identifications. For the purpose of this work, we assume that \(\lambda_z\) has no effect during that phase and the side projected area of the UAV remains the same during MRFT; since \({^\mathcal{B}}{v}{_i} \gg {^\mathcal{B}}{v}{_z} \), and \({^\mathcal{B}}{v}{_i} \approx {^\mathcal{W}}{v}{_i} \) for \(i \in \{x, y\}\).
\section{Experimental Results} \label{sec:res}
In this section, we provide experimental data that showcases the peformance of the proposed filter. The platform for the experiments used was a Quanser Qdrone UAV with an Intel Atom x7-Z8750 processor, a BMI160 6-DOF IMU, an Intel AC 8620 Wifi module, and an Optitrack MoCap system. All the filter processing was done in real-time on-board the UAV. The codes were developed in MATLAB Simulink environment, and a C++ code was generated using Simulink Coder to run on the UAV.

This section is split into three parts. First, we demonstrate the use of DNN-MRFT to identify the parameters of the SISO models presented in sec. \ref{sec:linmodels}. Then we examine the proposed model along with the identified parameters, by using a figure-eight maneuver to showcase the capabilities of the proposed filtering scheme. Finally, a ramp test is carried out to show how incorporating filtered acceleration can enhance controller tracking. The aforementioned tests assumes a calibrated accelerometer, i.e. \({^\mathcal{S} _\mathcal{B}}{R}{}=\mathcal{I}_{3 \times 3}\) with proper scaling adjustment of the accelerometer gains. The calibration was carried out in accordance with the method outlined in \cite{6pointtumble}. A recording of all the experiments in this section is available in \cite{wahbah2021_expvid}.
\subsection{Online Parameter Estimation Using DNN-MRFT}
We first utilize the DNN-MRFT framework to identify the UAV model parameters as described in sec. \ref{sec:linmodels}. At hover, we successively replace the attitude and position controllers by the MRFT algorithm to induce the oscillations needed. The identified parameters obtained from these tests are shown in Table \ref{tab:modparam}. The attitude loops were estimated to have similar dynamics, as the drone is symmetrical in design. The slight difference in process gain can be attributed to a slight weight distribution imbalance. For the altitude loop, the estimated process had considerably higher delay compared to the attitude, this is expected as a MoCap system have larger delays compared to the on-board inertial measurements. Finally, for the lateral loops, the test was carried out with a small MRFT relay height to reduce the attitude angles during the test. The maximum angular amplitudes recorded during the MRFT phase for the lateral loops \(b_x\) and \(b_y\) were 0.1042 rad, and 0.0972 rad, respectively. Thus it was assumed that the damping constants \(T_{\lambda_x}\) and \(T_{\lambda_y}\) were due to lateral motion only.
\begin{figure*} [t]
\centering
  \includegraphics[width = \textwidth]{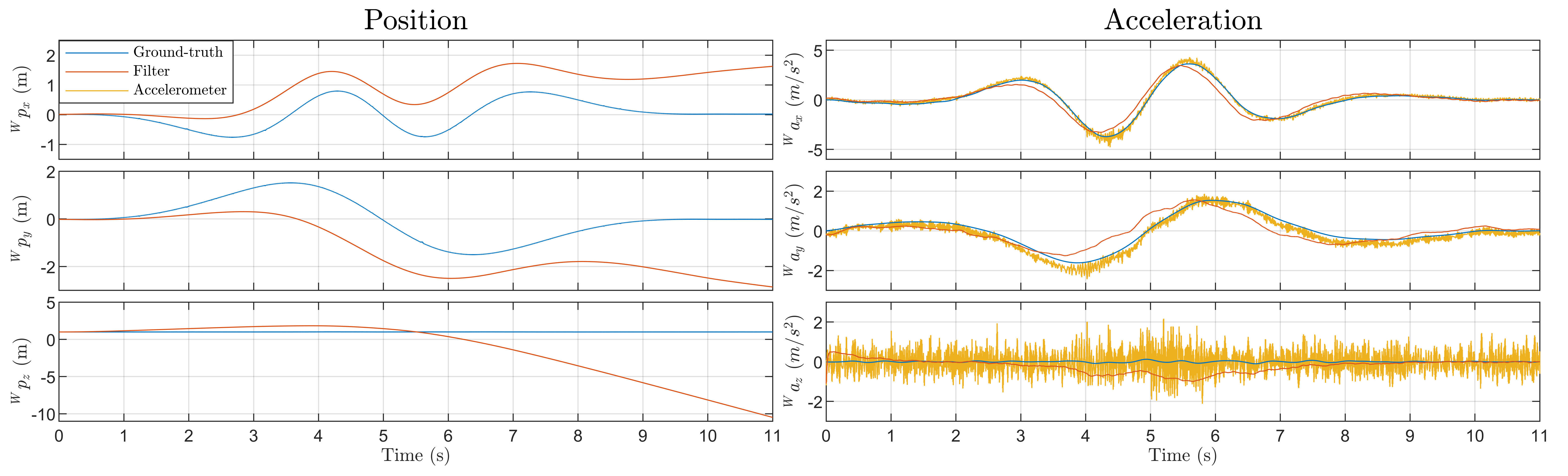}
\caption{Proposed translational filter open-loop performance}
\label{fig:ol}
\end{figure*}

Before examining the online performance of the filters, it would be beneficial to see how well the identified model can estimate the states when running in open-loop. The open-loop performance can be meaningfully investigated for translational dynamics, as in some practical scenarios position measurements might be absent. Specifically, we test the TDEKF estimator performance by neither updating position nor acceleration measurements. Fig \ref{fig:ol} shows the estimated position, and acceleration, using the TDEKF model shown in \ref{fig:TDEKFmodel} when performing a figure-eight trajectory. The estimated accelerations follow the ground truth trend, which indicates that the predicted model is close to the actual system. However due to sudden drop in battery levels and nonlinear asymmetries, we can see that during large accelerations along \(\vb*{w_x}\) and \(\vb*{w_y}\) the estimate along \(\vb*{w_z}\) diverges significantly from the ground truth. These differences in accelerations would accumulate and cause a drift in velocity and position.
\begin{table}[h]
\begin{center}
\caption{Identified model parameters}
\begin{tabular}{ |l|c| }
 \hline
 Model Parameter & Estimated Value \\
 \hline
 \(\tau_{b_x}\) & 0.0009 s \\
 \(K_{{eq_{b_x}}}\) & 72.6454 \\ 
 \(T_{prop_{b_x}}\) & 0.064 s \\ 
 \(T_{\gamma_x}\) & 0.2494 s \\ 
 \(\tau_{b_y}\) & 0.0009 s \\
 \(K_{{eq_{b_y}}}\) & 75.846 \\ 
 \(T_{prop_{b_y}}\) & 0.064 s \\ 
 \(T_{\gamma_y}\) & 0.2494 s \\ 
 \(\tau_{z}\) & 0.053 s \\
 \(K_{{eq}_z}\) & 0.2949 \\
 \(T_{prop}\) & 0.0177 s \\ 
 \(T_{\lambda_x}\) & 1.1629 s \\ 
 \(T_{\lambda_y}\) & 1.1629 s \\ 
 \(T_{\lambda_z}\) & 0.5793 s \\ 
 \hline
\end{tabular}
\label{tab:modparam}
\end{center}    
\end{table}
\subsection{Figure-Eight Maneuver}
\begin{figure*} [ht]
\begin{subfigure}{.5\textwidth}
  \centering
  \includegraphics[width=\linewidth]{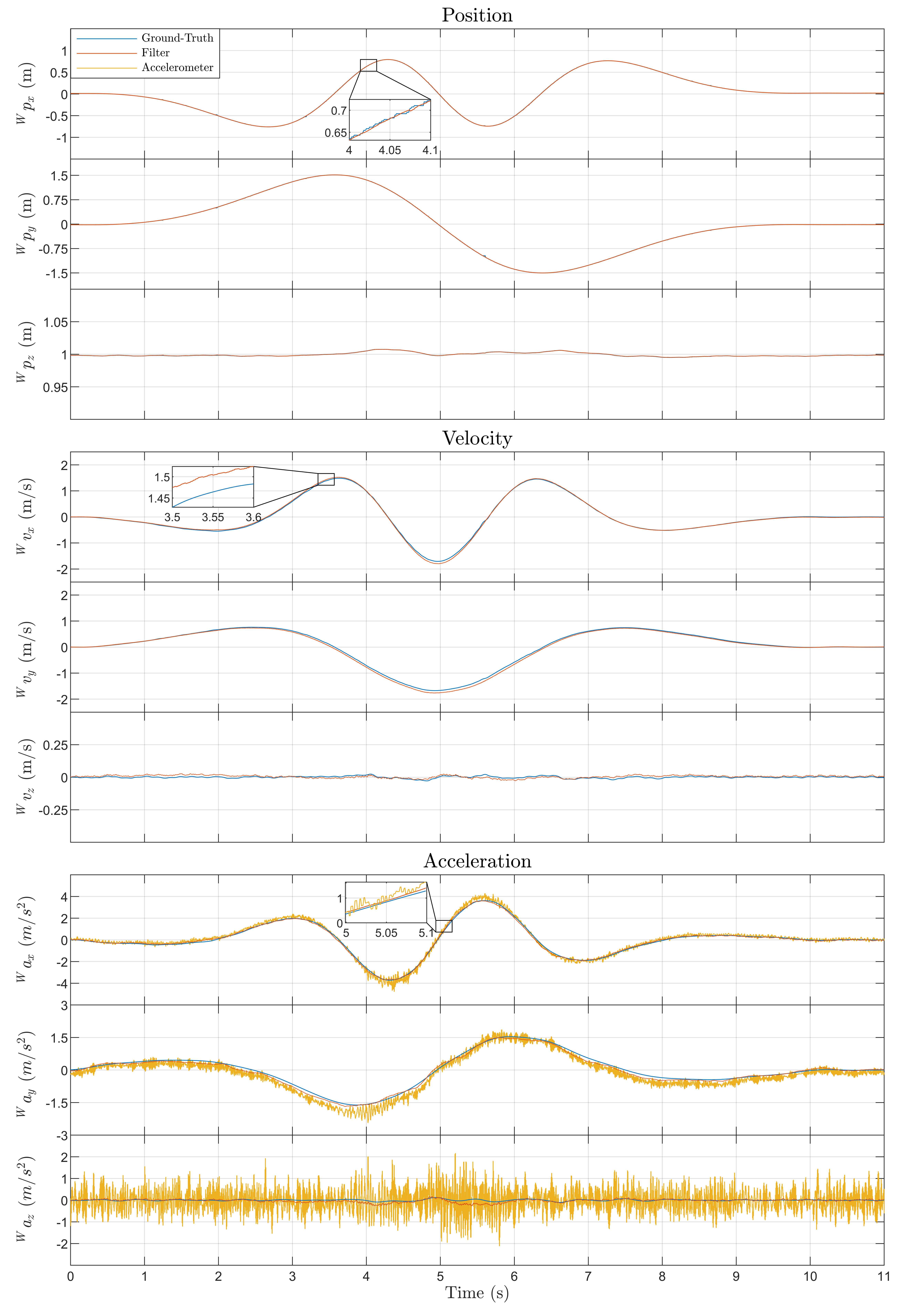}  
  \caption{With 300 Hz position update}
  \label{fig:300compact}
\end{subfigure}
\begin{subfigure}{.5\textwidth}
  \centering
  \includegraphics[width=\linewidth]{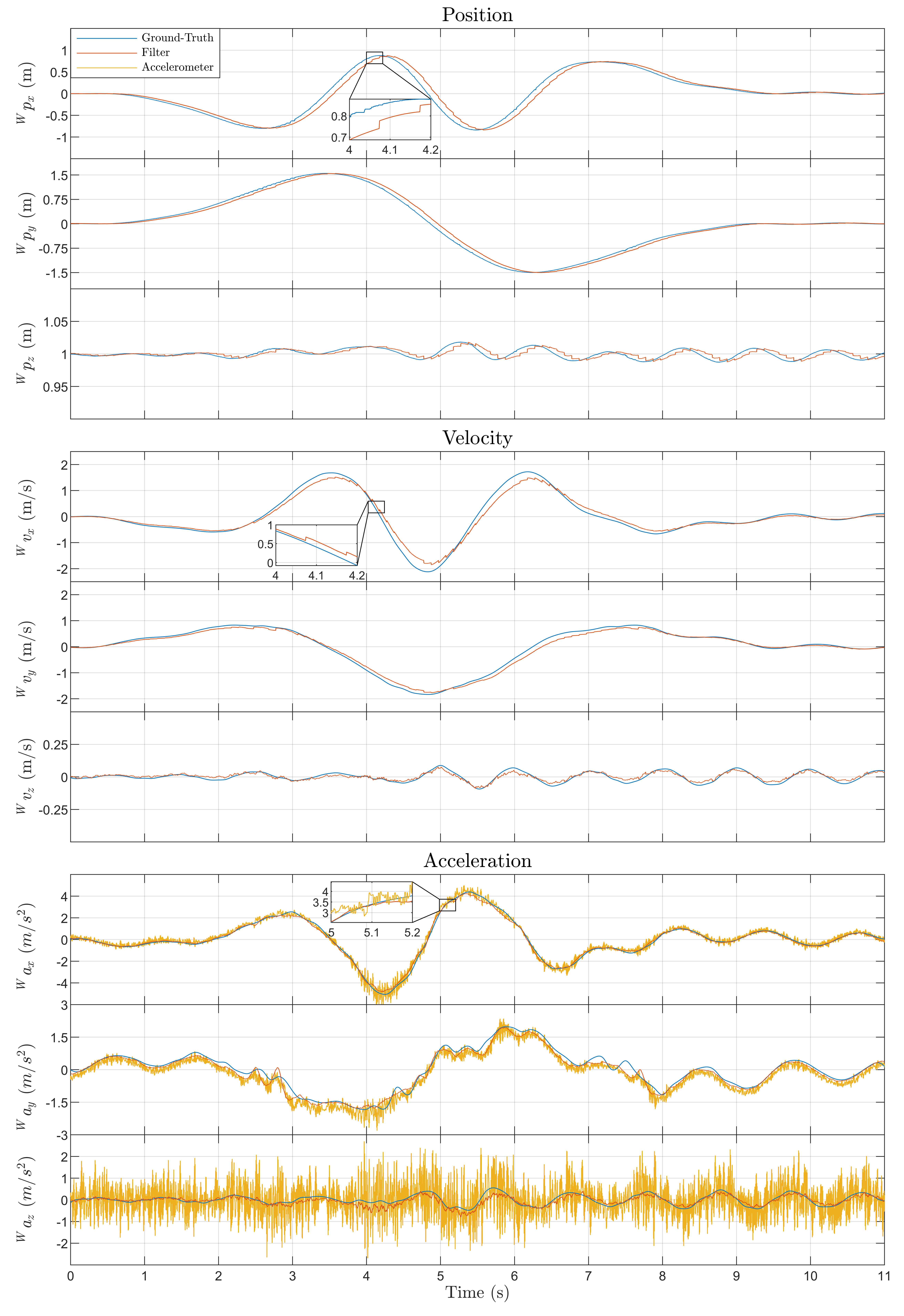}  
  \caption{With 10 Hz position update}
  \label{fig:10compact}
\end{subfigure}
\caption{Proposed translational filter estimates}
\label{fig:tdekf_est}
\end{figure*}

To assess the performance of the proposed filtering scheme, along with the identified model, we perform two figure-eight maneuvers with two different update rates of the position measurements. In the first figure-eight maneuver, the filter provides the estimated position, orientation, and inertial and rotational velocities for feedback control at a rate of 300Hz. Then, we repeat the previous test with a down sampled position measurement of 10Hz. We chose the 10Hz update rate as most commercially available RTK or UWB positioning systems provide position measurements at least at 10Hz. RTK and UWB are the popular choices for UAV's operated outdoors, where a MoCap system is not always feasible. Note that since the IMU measurements and the prediction algorithms are running at 1kHz, down sampling the position measurements will only affect the TDEKF performance, and hence such comparative assessment does not hold for the RDEKF.
\begin{figure}
\centering
  \includegraphics[width = \linewidth]{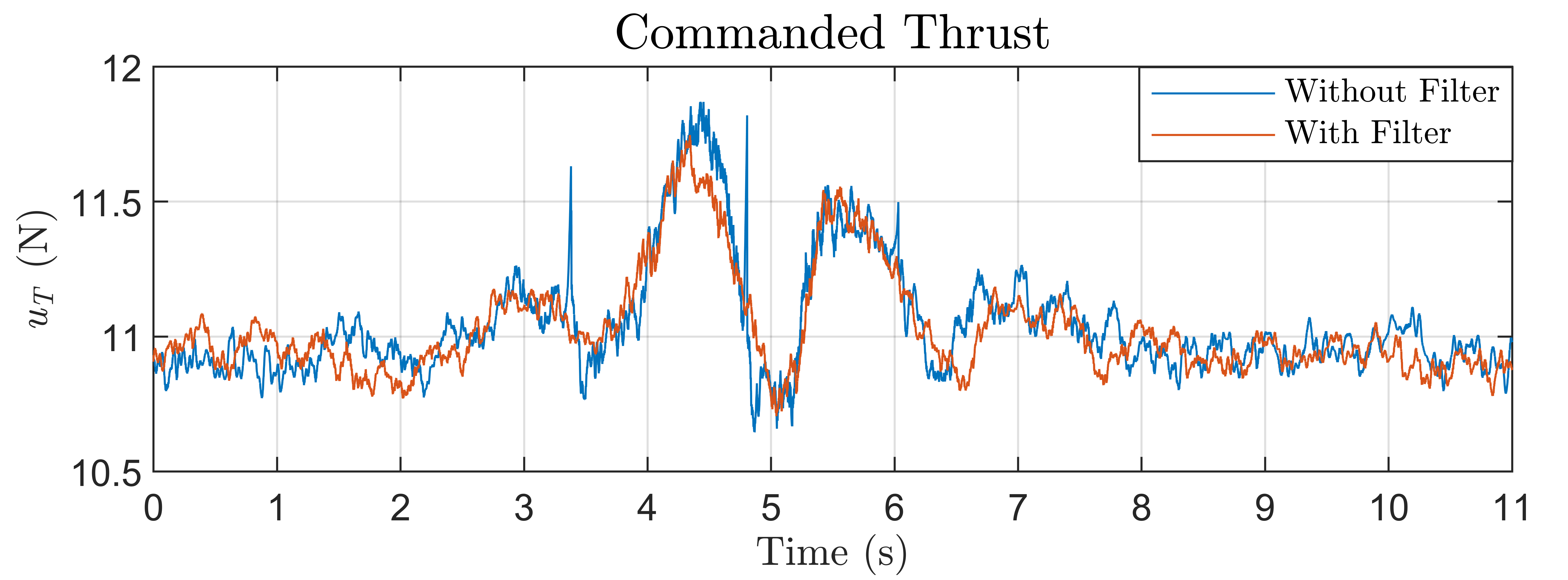}
\caption{Controller command reduction when using the proposed filtering framework}
\label{fig:u_reduce}
\end{figure}
Fig. \ref{fig:300compact} shows the position estimation performance of the TDEKF with 300Hz measurements update when performing a ten seconds figure-eight trajectory. In this case we can see that the estimates closely match the ground truth and perform predictions between position updates to provide estimates position and velocity estimates at 1kHz. The TDEKF smoothed the high rate MoCap position measurements which resulted in a notable reduction of the controller energy leading to a smoother flight. Fig. \ref{fig:u_reduce} shows the thrust commands produced by the controller when operated using the raw measurements in one case, and the filtered estimates in another case. It can be noticed that the high frequency variations in \(u_T\) reduced when using the filter. The reduction in controller action was around \(6.6 \%\), increasing signal to noise ratio (SNR) from \(15.73 \; dB\) to \(20.96 \; dB\), high frequencies were all assumed to be noises. When providing measurement updates at 10Hz, we observed that the velocity and position estimates lags behind the ground truth possibly due to the added delay introduced by the down sampling, however, we can see that it still provides smooth and stable estimates of position and velocity that are suitable for feedback control (refer to Fig. \ref{fig:10compact}).  

In order to systematically assess the difference in filter performance for the two different measurement update rate cases we defined a ground truth quantity and proposed two quantitative metrics. For all results the ground truth is obtained from the post-processing of the MoCap measurements. Post-processing improves the accuracy of the ground truth due to non-causality. In the first quantitative metric we computed the Root Mean Square Error (RMSE) between the filter estimates and the ground truth. The RMSE is calculated using the following equation:
\begin{equation}
    RMSE = \sqrt{\dfrac{\sum _{i=1} ^N (\mathbf{x}-\mathbf{x^{ref}})_i^2}{N}}
\end{equation}
where \(\mathbf{x}\) is a vector of measured values, \(\mathbf{x^{ref}}\) is a vector of reference values, and $N$ is the number of the measurement samples in the trajectory. The results presented in Table \ref{tab:rmse} show that the RMSE figures increase in the 10Hz case, but the errors remain small for the practical interest. The increase in the error figures was smallest for the acceleration states, then the velocity states, and was largest for the position states. Such distribution of the increase of the error figures can be changed by the tuning of process noise \(Q\) matrix. We found that with the proposed filter structure, the RMSE results slightly change for a wide range of \(Q\) matrix tuning, which makes the implementation of the TDEKF and RDEKF straightforward and easily applicable to a wide range of UAV designs. It can be difficult to compare our results with those reported in literature, due to variations in hardware and experimental setup. However using RMSE as a metric should give an indication of our filter performance, as it normalizes the error over time. Hence we will compare the performance using RMSE. Table \ref{tab:comprison} shows a comparison between our achieved RMSE and those from \cite{Yong2019}. The authors in \cite{Yong2019} carried out an experiment using a Global Navigation Satellite System (GNSS) following a rectangular path to asses their dynamics based filter performance. The position RMSE's reported were \(0.3109 \; m\), \(0.5895 \; m\), and \(1.3527 \; m\), for \({^\mathcal{W}}p_x\), \({^\mathcal{W}}p_y\), and \({^\mathcal{W}}p_z\) respectively. Velocity estimates were also evaluated, and the filter achieved RMSE's of \(0.0803 \; m/s\), \(0.0757 \; m/s\), and \(0.0838 \; m/s\) for \({^\mathcal{W}}v_x\), \({^\mathcal{W}}v_y\), and \({^\mathcal{W}}v_z\) respectively. From Table \ref{tab:comprison} we can see that our proposed approach achieved significantly better position estimates, albeit using a MoCap. MoCap have higher position accuracy than the GNSS, however both report high precision measurements, and were running at 10 Hz. Additionally, the GNSS provides piston and accurate velocity measurements, both of which were used to correct the estimates. Whereas the MoCap only provides position measurement. To that effect, the velocity estimates reported by \cite{Yong2019} slightly out-preformed ours. Due to the novelty of including the acceleration as a state, we can not compare with other dynamic based filters from literature. Authors in \cite{Hamandi2020} did however propose a regression based notch filter for removing the noise form accelerometer measurements in UAVs. The performance of the filter was not analyzed quantitatively, but the authors claimed that the filter removed most of the noises, without introducing a lag or attenuation in the signal. Similarly, our proposed filter did not introduce notable lag in estimating the acceleration, even during the 10 Hz position measurements. Additionally the filter presented in \cite{Hamandi2020} required collecting data and tuning the filter parameter for every UAV setup used. Whereas our approach is fully automated.

In the second quantitative metric, we show the drop in control performance caused by down sampling the position measurements. This drop in the control performance is quantified by the RMSE and contouring errors (CE) between the ground truth and the reference trajectory. The following equation defines the average CE:
\begin{equation}
    CE =\sum_{i=1}^N \min \left( \sqrt{(\mathbf{x_i} - \mathbf{x^{ref}})^2} \right)
\end{equation}
The results for the drop in control performance are reported in Table \ref{tab:cont_det}. The results show that the CE increase due to down sampling is much lower than the increase in the RMSE, which implies that the 10Hz position estimates remain still close to the ground truth, but are considerably lagged. These results are promising for trajectory tracking applications where only low update rate position sensors are available, and lag in trajectory execution is tolerated.

\begin{figure}
\centering
  \includegraphics[width = \linewidth]{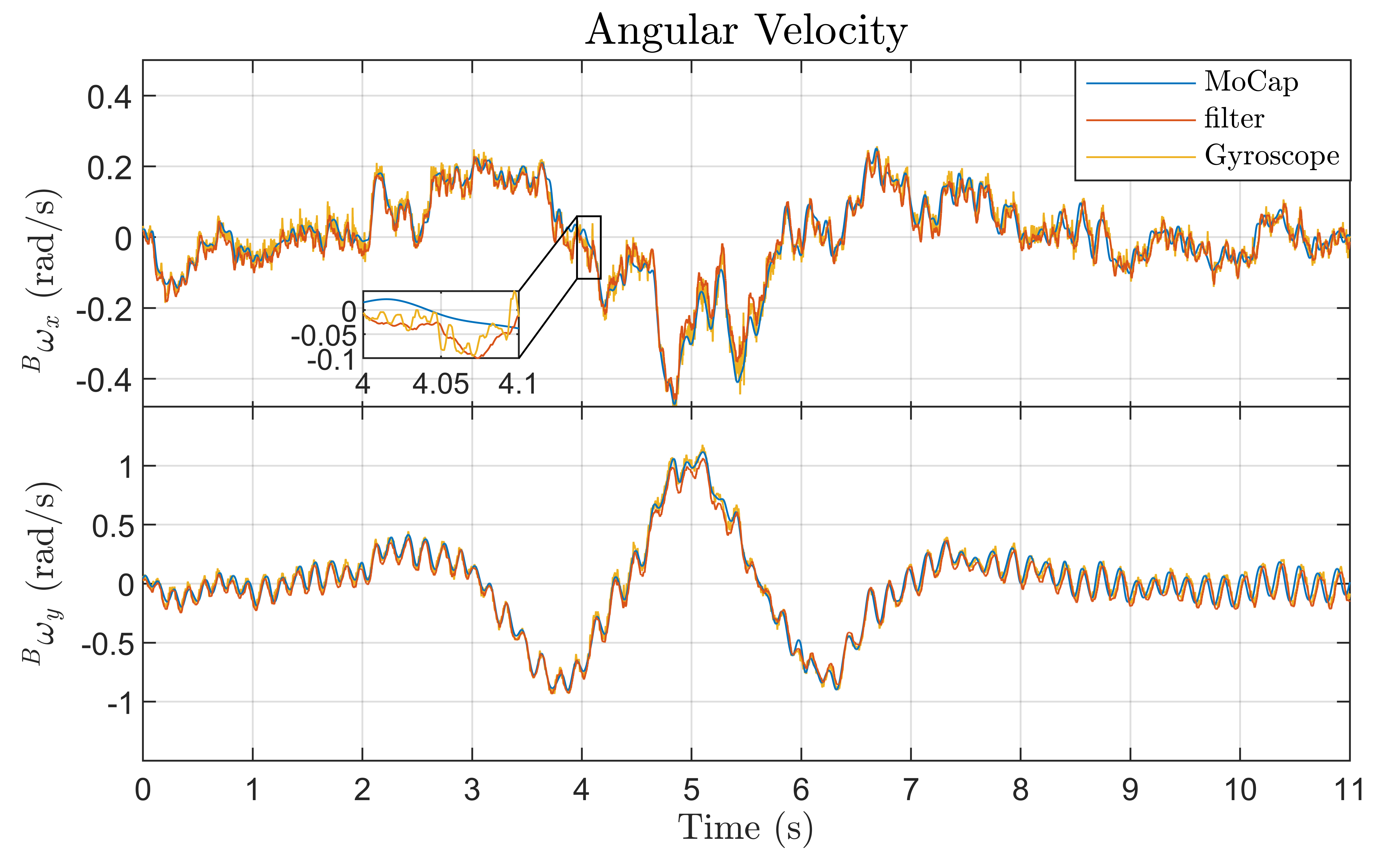}
\caption{Proposed rotational filter angular velocity estimates in the 300 Hz position update test}
\label{fig:omega300}
\end{figure}

The RDEKF angular velocity estimates are provided for the 300Hz case in Fig. \ref{fig:omega300}, with their respective RMSE shown in Table \ref{tab:rmse}. The oscillations with a frequency close to 7Hz are due to amplified low frequency body vibrations. The filter successfully rejects all higher frequencies compared to the raw gyroscope measurements without lagging 
\begin{table}[]
\begin{center}
\caption{Estimates RMSE figures measuring deviations between the filters' estimates and the ground truth}
    \begin{tabular}{|c|c|c|c|}
        \hline
        \multicolumn{2}{|c|}{\multirow{2}{*}{State}}&\multicolumn{2}{c|}{RMSE}\\
        \cline{3-4}
        \multicolumn{2}{|c|}{}&300 Hz & 10 Hz\\
        \hline
        \multirow{3}{*}{Position} & \({^\mathcal{W}}p_x\) & 0.0027 m & 0.0903 m \\
                                  & \({^\mathcal{W}}p_y\) & 0.0029 m & 0.0764 m \\
                                  & \({^\mathcal{W}}p_z\) & 0.0001 m & 0.0044 m \\
        \hline
        \multirow{3}{*}{Velocity} & \({^\mathcal{W}}v_x\) & 0.0297 m/s & 0.1573 m/s \\
                                  & \({^\mathcal{W}}v_y\) & 0.0475 m/s & 0.0919 m/s \\
                                  & \({^\mathcal{W}}v_z\) & 0.0114 m/s & 0.0147 m/s \\
        \hline
        \multirow{3}{*}{Acceleration} & \({^\mathcal{W}}a_x\) & 0.0809 \(m/s^2\) & 0.1542 \(m/s^2\) \\
                                      & \({^\mathcal{W}}a_y\) & 0.0812 \(m/s^2\) & 0.1769 \(m/s^2\) \\
                                      & \({^\mathcal{W}}a_z\) & 0.0638 \(m/s^2\) & 0.1286 \(m/s^2\) \\
        \hline
        \multirow{2}{*}{Angle} & \(\theta\) & 0.0088 rad & \cellcolor{Gray} \\
                               & \(\phi\) & 0.0060 rad & \cellcolor{Gray} \\
        \hline
        \multirow{2}{*}{Rotational Speed} & \({^\mathcal{B}}\omega_x\) & 0.0296 rad/s & \cellcolor{Gray} \\
                                          & \({^\mathcal{B}}\omega_y\) & 0.0621 rad/s & \cellcolor{Gray} \\
        \hline
    \end{tabular}
\label{tab:rmse}
\end{center}    
\end{table}
\begin{table}[]
    \begin{center}
        \caption{Figure-eight tracking RMSE \& CE measuring the deviation between the ground truth and the reference trajectory.}
        \begin{tabular}{|c|c|c|c|}
            \hline
            \multicolumn{2}{|c|}{\multirow{2}{*}{State}}&\multicolumn{2}{c|}{RMSE}\\
            \cline{3-4}
            \multicolumn{2}{|c|}{}&300 Hz & 10 Hz\\
            \hline
            \multirow{3}{*}{Position} & x & 0.0204 m & 0.1197 m \\
                                      & y & 0.0127 m & 0.0852 m \\
                                      & z & 0.0030 m & 0.0069 m \\
            \hlinewd{1pt}
                    \multicolumn{2}{|c|}{\multirow{2}{*}{State}}&\multicolumn{2}{c|}{CE}\\
            \cline{3-4}
            \multicolumn{2}{|c|}{}&300 Hz & 10 Hz\\
            \hline
            \multicolumn{2}{|c|}{Position} & 0.0097 m & 0.0273 m \\
            \hline
        \end{tabular}
        \label{tab:cont_det}
    \end{center}
\end{table}
\begin{table}[]
\begin{center}
\caption{Comparison between our proposed approach and 3D-DMAN-UKF with GNSS from \cite{Yong2019}}
\resizebox{\columnwidth}{!}{
    \begin{tabular}{|c|c|c|c|}
        \hline
        \multicolumn{2}{|c|}{\multirow{2}{*}{State}}&\multicolumn{2}{c|}{RMSE}\\
        \cline{3-4}
        \multicolumn{2}{|c|}{}& TDEKF with 10 Hz MoCap & 3D-DMAN-UKF with GNSS\\
        \hline
        \multirow{3}{*}{Position} & \({^\mathcal{W}}p_x\) & 0.0903 m & 0.3109 m \\
                                  & \({^\mathcal{W}}p_y\) & 0.0764 m & 0.5895 m \\
                                  & \({^\mathcal{W}}p_z\) & 0.0044 m & 1.3527 m \\
        \hline
        \multirow{3}{*}{Velocity} & \({^\mathcal{W}}v_x\) & 0.1573 m/s & 0.0803 m/s \\
                                  & \({^\mathcal{W}}v_y\) & 0.0919 m/s & 0.0757 m/s \\
                                  & \({^\mathcal{W}}v_z\) & 0.0147 m/s & 0.0838 m/s \\
        \hline
    \end{tabular}
    }
\label{tab:comprison}
\end{center}
\end{table}
\subsection{Position Ramp Test}
\begin{figure}
\centering
  \includegraphics[width = \linewidth]{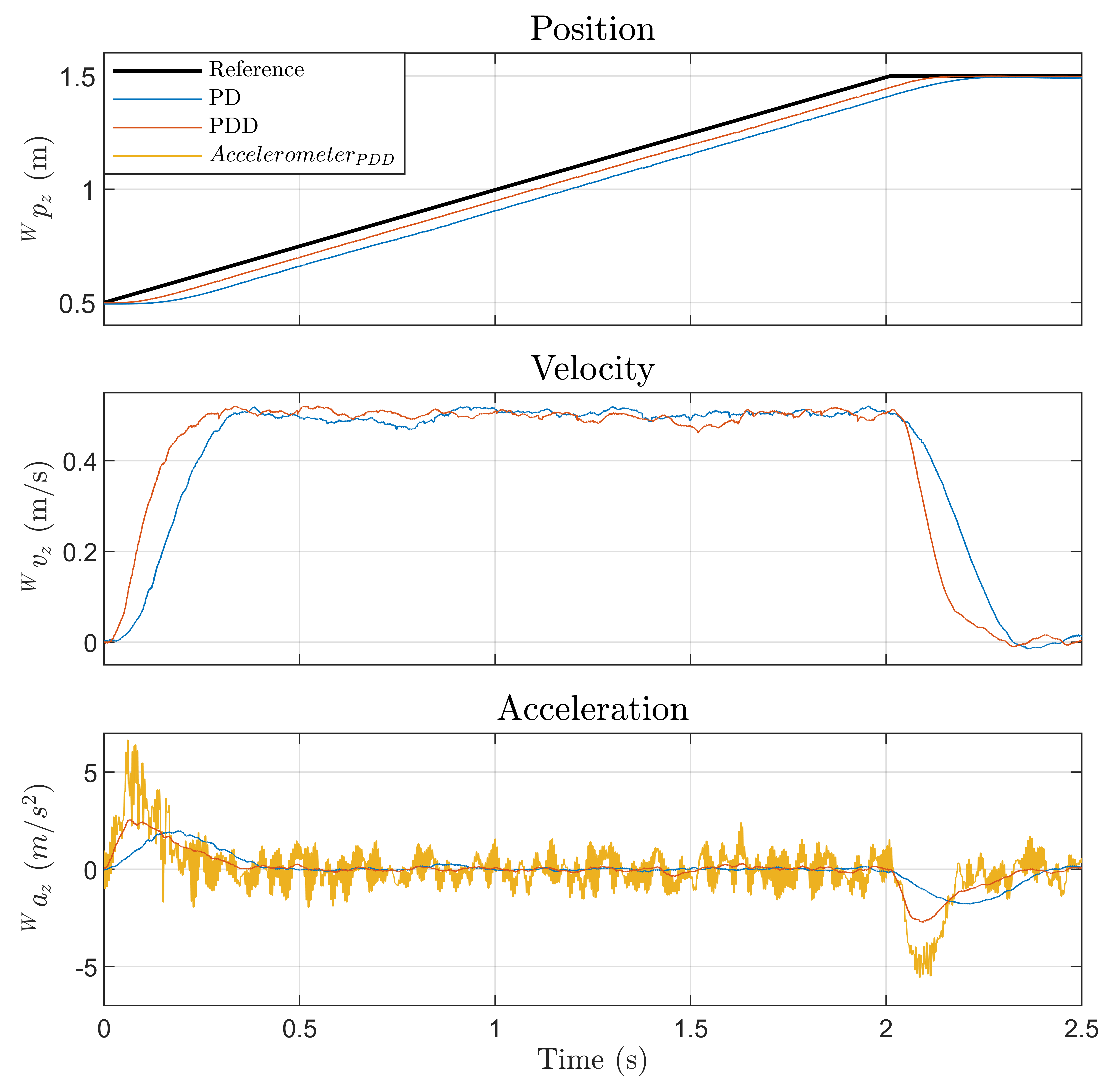}
\caption{Ramp tracking performance of PD and PDD controllers using the proposed translational filter estimates}
\label{fig:ramp}
\end{figure}

Although providing a control strategy that incorporates acceleration feedback is not in the scope of this article, a test that shows the advantages of using filtered acceleration in control is necessary to demonstrate the benefit of the proposed filtering scheme. In this test, we provide a ramp reference on \({^\mathcal{W}}p_z\) and compare the performance of a Proportional Derivative (PD) controller, to that of a Proportional Double Derivative (PDD) controller. The optimal PD and PDD values for the identified model were found offline. We then used the filter estimates with 300 Hz position measurements to track a ramp position reference with a slope of \(0.5 \; m/s \) for \(2\) seconds, thus moving the UAV a total of \(1 \; m\). As seen in Fig. \ref{fig:ramp} using a PDD controller provided faster response than a PD controller, as the RMSE  dropped from \(0.0791 \; m\) to \(0.0449 \; m\).
\section{Conclusion} \label{sec:conc}
This work further develops the controller tuning approach presented in \cite{ayyad2021tcst} for state estimation. We showed that the SISO models used for channel wise parameter identification is able to capture the dynamics of the UAV presented in literature. The decoupled approach for state estimation proposed in this paper was able to provide estimates for position, velocity, acceleration, and angular velocity. Although the decoupling of the filter requires accelerometer calibration, our approach is implemented online without any prior knowledge of UAV parameters, nor does it require additional measurements, such as motor speed. The efficacy of the proposed filtering scheme was validated experimentally by preforming a figure-eight maneuver using \(10 \; Hz\) position measurements. Such maneuvers are usually only feasible with high rate MoCap system. However our proposed filter was able to up-sample the position measurements to \(1 \; kHz\), thus executing the maneuver accurately and successfully. Additionally, our approach was also able to provide a smooth acceleration estimate, a feature that was not found in previous dynamic based filters for UAV's. Having knowledge about the inertial acceleration can increase control performance, as was demonstrated by a ramp tracking experiment. Future work can utilize our accurate estimator model to design algorithms that can detect external disturbances or sensors failure.

\section*{Acknowledgment}
The authors would like to thank Eng. Abulla Ayyad and Eng. Abdulaziz Alkayas for their technical assistance with the experiments.

\bibliographystyle{IEEEtran}
\bibliography{bib/ref.bib}

\begin{IEEEbiography}[{\includegraphics[width=1in,height=1.25in,clip,keepaspectratio]{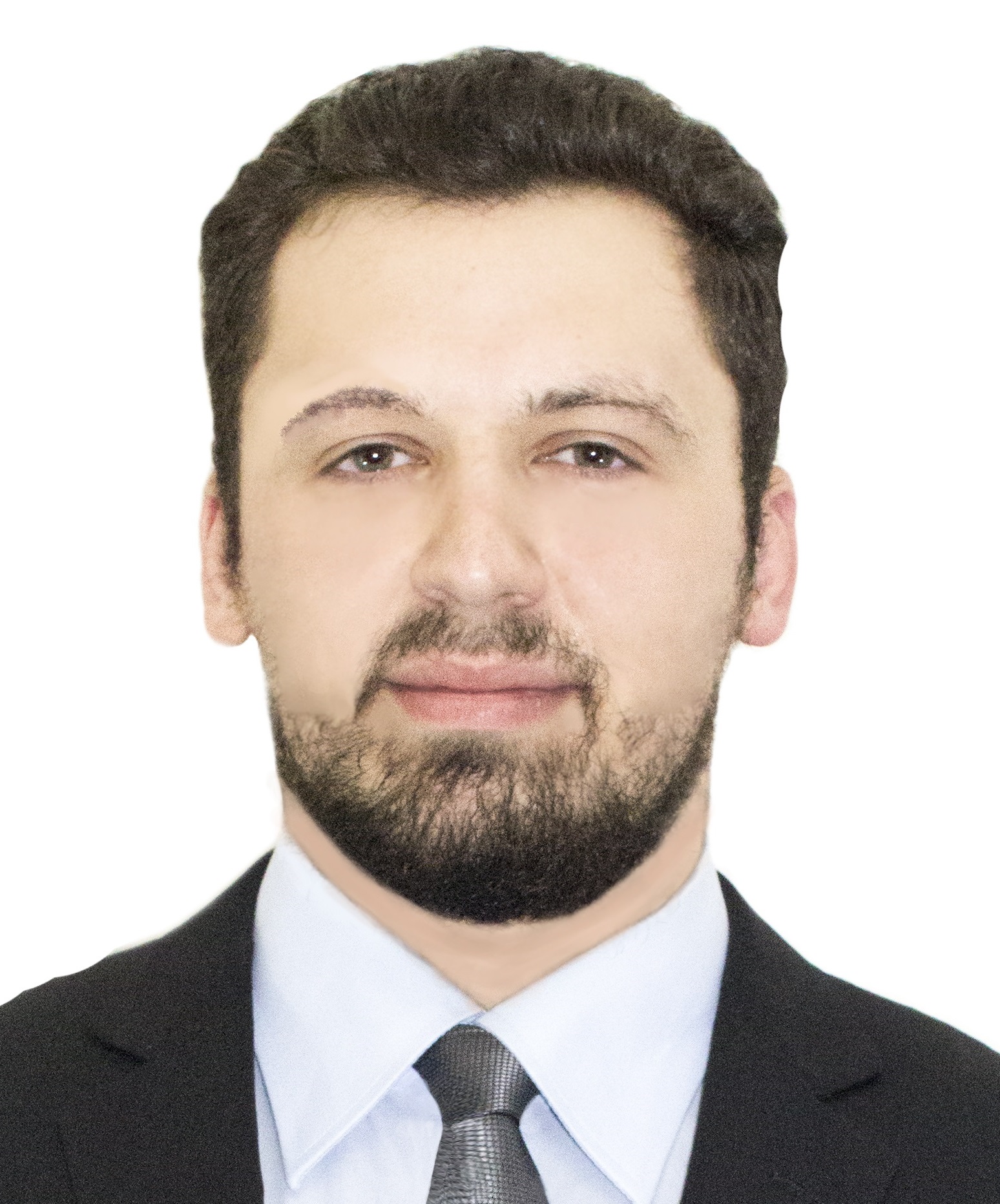}}]{Mohamad Wahabah} received his MSc. in Electrical Engineering from Khalifa University, Abu Dhabi, UAE, in 2018. He is currently a researcher with Khalifa University Center for Robotic Systems (KUCARS). His research areas include multisensor fusion, state estimation, and navigation in hazardous environments.
\end{IEEEbiography}
\begin{IEEEbiography}[{\includegraphics[width=1in,height=1.25in,clip,keepaspectratio]{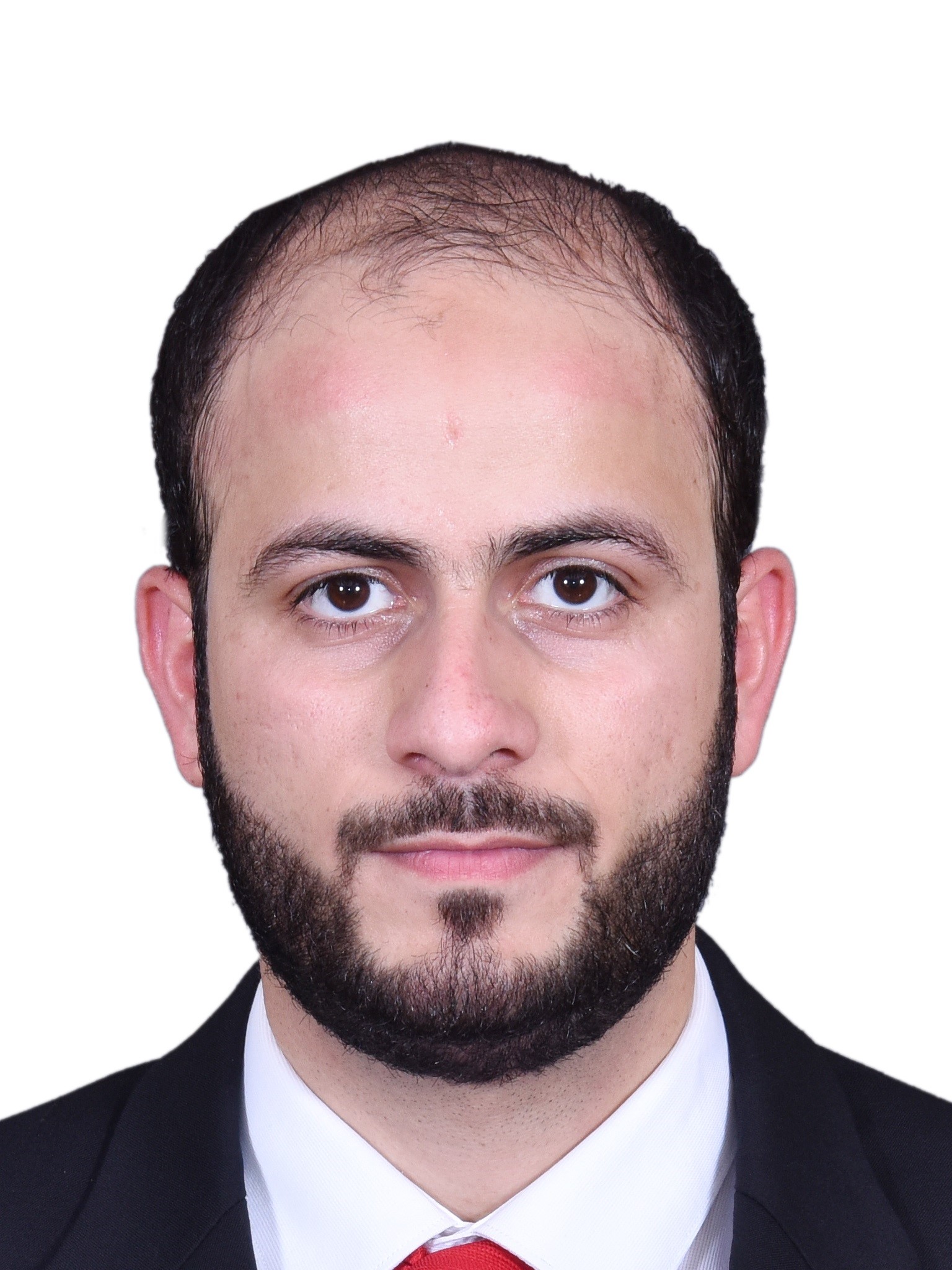}}]{Mohamad Chehadeh} received his MSc. in Electrical Engineering from Khalifa University, Abu Dhabi, UAE, in 2017. He is currently with Khalifa University Center for Autonomous Robotic Systems (KUCARS). His research interest is mainly focused on identification, perception, and control of complex dynamical systems utilizing the recent advancements in the field of AI.
\end{IEEEbiography}
\begin{IEEEbiography}[{\includegraphics[width=1in,height=1.25in,clip,keepaspectratio]{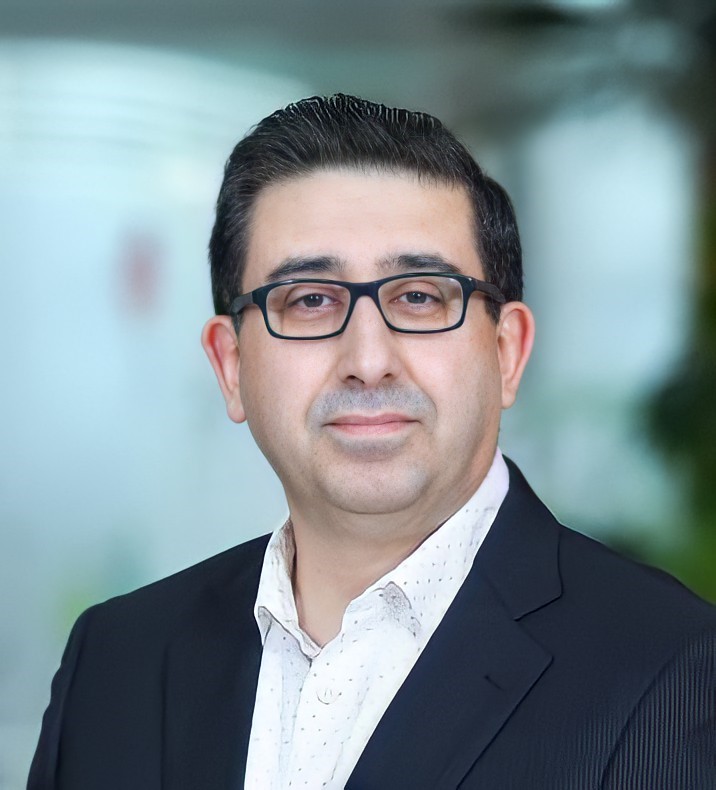}}]{Yahya Zweiri} (Member, IEEE) received the
Ph.D. degree from King's College London, in 2003. He is currently the School Director of the Research and Enterprise, Kingston University London, U.K. He is also an Associate Professor with the Department of Aerospace, Khalifa University, United Arab Emirates. He was involved in defense and security research projects in the last 20 years at the Defence Science and Technology Laboratory, King's College London, and the King Abdullah II Design and Development Bureau, Jordan. He has published over 100 refereed journal and conference papers and filed six patents in USA and U.K. in the unmanned systems field. His central research interests include interaction dynamics between unmanned systems and unknown environments by means of deep learning, machine intelligence, constrained optimization, and advanced control.
\end{IEEEbiography}

\EOD
\end{document}